\documentclass[final,5p,times,twocolumn]{elsarticle}

\usepackage{amsmath,amssymb,amsfonts}
\usepackage{algorithmic}
\usepackage{graphicx}
\usepackage{algorithm,algorithmic}

\usepackage{textcomp}

\usepackage{comment}
\usepackage{subcaption}
\usepackage{siunitx}
\usepackage{interval}
\usepackage[colorlinks=true,linkcolor=blue]{hyperref} 
\usepackage{booktabs}
\usepackage{multirow} 
\usepackage{makecell} 
\usepackage[acronym]{glossaries} 
\usepackage[table]{xcolor}
\usepackage{microtype} 
\usepackage{etoolbox}
\usepackage{enumitem}

\glsdisablehyper 

\newacronym{ai}{AI}{artificial intelligence}
\newacronym{auprc}{AUPRC}{area under the precision-recall curve}
\newacronym{auroc}{AUROC}{area under the receiver operating characteristic curve}
\newacronym{atc}{ATC}{Anatomical Therapeutic Chemical}
\newacronym{bert}{BERT}{bidirectional encoder representations from Transformers}
\newacronym{bmi}{BMI}{body mass index}
\newacronym{chv}{CHV}{cross-hospital validation}
\newacronym{ci}{CI}{confidence interval}
\newacronym{dl}{DL}{deep learning}
\newacronym{dx}{DX}{diagnoses} 
\newacronym{ehr}{EHR}{electronic health record}
\newacronym{gmacs}{GMACs}{giga multiply-add operations}
\newacronym{hf}{HF}{heart failure}
\newacronym{icd}{ICD-10}{International Classification of Diseases 10th revision}
\newacronym{iqr}{IQR}{interquartile range}
\newacronym{lab}{LAB}{laboratories}
\newacronym{lstm}{LSTM}{long short-term memory}
\newacronym{med}{MED}{medications}
\newacronym{mlm}{MLM}{masked language modeling}
\newacronym{ntp}{NTP}{next token prediction}
\newacronym{plm}{PLM}{permutation language modeling}
\newacronym{pro}{PRO}{procedures}
\newacronym{rnn}{RNN}{recurrent neural network}
\newacronym{rq}{RQ}{research question}
\newacronym{shv}{SHV}{single-hospital validation}
\newacronym{ssm}{SSM}{state space model}
\newacronym{tgds}{TGDS-HF}{trajectory-guided discharge stratification for heart failure}
\newacronym{vit}{VIT}{vital signs}
\newacronym{xgb}{XGBoost}{eXtreme gradient boosting machine}

\newcommand{\vocone}{$V_{b=5,i=3}$}
\newcommand{\voctwo}{$V_{b=10,i=3}$}
\newcommand{\vocthree}{$V_{b=5,i=4}$}
\newcommand{\vocfour}{$V_{b=10,i=4}$}

\newcommand{\histtzero}{$H_{t=0}$} 
\newcommand{\histtone}{$H_{t\leq 1 \mathrm{y}}$} 
\newcommand{\histtthree}{$H_{t\leq 3 \mathrm{y}}$} 
\newcommand{\histcontext}{$H_{t\leq C}$} 
\newcommand{\histwinone}{$H_{t\leq C,w=1\mathrm{d}}$} 
\newcommand{\histwintwo}{$H_{t\leq C,w=2\mathrm{d}}$} 

\newcommand{\ci}[1]{{\scriptsize\textcolor{gray}{#1}}}

\AtBeginEnvironment{tabular}{\footnotesize}
\AtBeginEnvironment{tabular*}{\footnotesize}

\begin{document}

\begin{frontmatter}

\title{Trajectory-guided discharge stratification for heart failure using short-context electronic health record sequence modeling}

\author[label1]{Falk Dippel}
\author[label2,label7]{Yinan Yu\corref{cor1}}
\author[label3,label4]{Annika Rosengren}
\author[label3,label4]{Martin Lindgren}
\author[label4,label5]{Christina E. Lundberg}
\author[label2]{Erik Aerts}
\author[label6]{Martin Adiels} 
\author[label3,label4,label7]{Helen Sjöland}

\affiliation[label1]{
    organization={Sahlgrenska University Hospital},
    addressline={Diagnosvägen 11},
    city={Gothenburg},
    postcode={416 85},
    country={Sweden}
}

\affiliation[label2]{
    organization={Department of Computer Science and Engineering, Chalmers University of Technology and University of Gothenburg},
    addressline={Rännvägen 6B},
    city={Gothenburg},
    postcode={412 96},
    country={Sweden}
}

\affiliation[label3]{
    organization={Department of Molecular and Clinical Medicine, Sahlgrenska Academy, University of Gothenburg},
    addressline={Bruna stråket 16},
    city={Gothenburg},
    postcode={413 45},
    country={Sweden}
}

\affiliation[label4]{
    organization={Department of Medicine, Geriatrics and Emergency Medicine, Sahlgrenska University Hospital},
    addressline={Diagnosvägen 11},
    city={Gothenburg},
    postcode={416 85},
    country={Sweden}
}

\affiliation[label5]{
    organization={Department of Food and Nutrition, and Sport Science, Faculty of Education, University of Gothenburg},
    addressline={Box 300},
    city={Gothenburg},
    postcode={405 30},
    country={Sweden}
}

\affiliation[label6]{
    organization={School of Public Health and Community Medicine, Institute of Medicine, University of Gothenburg},
    addressline={Box 453},
    city={Gothenburg},
    postcode={405 30},
    country={Sweden}
}

\affiliation[label7]{
    organization={Center of Digital Health, Sahlgrenska University Hospital},
    city={Gothenburg},
    postcode={413 45},
    country={Sweden}
}

\cortext[cor1]{Corresponding author: \url{yinan@chalmers.se}}

\begin{abstract} 
    \textbf{Purpose:} \Gls{hf} discharge planning depends on identifying patients at risk of deterioration or death, yet accurate prediction from routinely collected \glspl{ehr} remains challenging. \textbf{Methods:} We develop \gls{tgds}, a methodology that reads the patient in-hospital trajectory of diagnoses, vital signs, laboratories, medications, and procedures end-to-end with a compact short-context autoregressive Transformer, and uses it to stratify one-year risks of clinical instability (a rehospitalization phenotype) or mortality for discharge care. We instantiate \gls{tgds} on a Swedish \gls{hf} cohort ($N=42{,}820$) to predict one-year clinical instability or mortality at the initial \gls{hf} diagnosis in-hospital. \Gls{tgds} has three components: category-level tokenization, recency-weighted temporal representation, and sequence model configuration. We run ablations on these components to show the effectiveness of \gls{tgds}. \textbf{Results:} Against traditional \gls{xgb} and \gls{bert}-based \gls{ehr} sequence-modeling baselines, \gls{tgds} (Llama backbone) achieved \glspl{auprc} with 95\% confidence intervals of \mbox{0.555 (0.535--0.575)} and \mbox{0.574 (0.550--0.599)} across the two tasks at the method default, with robust calibration. A task-specific refinement using daily aggregation of repeated continuous events improves the mortality task to \mbox{0.582 (0.558--0.608)}. Further, \Gls{tgds} maintains strong performance under reduced clinical concept availability and limited training data. \textbf{Conclusion:} Combined predictions of instability and mortality from \Gls{tgds} may support personalized  discharge planning, ranging from follow-up in primary care to specialist-led management and, when appropriate, palliative care.
\end{abstract}

\begin{keyword}
Electronic health record; Transformer; Mamba; Heart failure; Mortality prediction; Readmission prediction
\end{keyword}

\journal{Journal}

\end{frontmatter} 

\glsresetall 

\section{Introduction} \label{sec:introduction}
\Gls{hf} is a chronic, progressive condition with frequent rehospitalizations  and high mortality  \cite{groenewegen2020epidemiology}. During a \gls{hf} hospitalization, clinicians face a critical decision: what level of follow-up does this patient need? Patients identified as high risk for clinical instability (a phenotype of readmission to hospital) may benefit from specialized monitoring, whereas those at high risk of mortality may require tailored intensive care programs or palliative care referral. Conversely, patients at low risk may be appropriately managed through routine primary care follow-up, including annual clinical assessments. Such decisions are crucial for both appropriate care delivery and healthcare cost control. Currently, this risk stratification relies largely on clinical judgment, since alternatives such as simple scoring systems tend to perform poorly for decision making \cite{canepa2018performance, oscar2020risk}. Accurate, individualized predictions of one-year clinical instability and mortality using \gls{ai} systems could transform discharge planning into a structured, data-driven process and decision support. However, unreliable predictions place considerable demands and cost on the healthcare system and lead to suboptimal patient care decisions, motivating the development of precise \gls{ai} systems.

Patient health information is commonly stored and collected in longitudinal \glspl{ehr} by healthcare providers, capturing a patient trajectory through irregular clinical visits including hospitalizations. An \gls{ehr} comprises structured data, e.g. recorded vital signs or assessed diagnoses, and unstructured data, e.g. clinical text notes or medical images \cite{hayrinen2008definition}. From a signal processing perspective, \glspl{ehr} can be understood as multimodal, irregularly sampled time series \cite{yadav2018mining}. In recent years, the application of \gls{ai} on \glspl{ehr} has gained significant research interest due to the growing availability of \glspl{ehr} and advances in \gls{ai} \cite{kim2019evolving, juhn2020artificial}. The field of natural language processing, in particular, has highlighted its efficiency to learn patient representations from \glspl{ehr} \cite{juhn2020artificial, steinberg2021language, wornow2023shaky}. Transformers \cite{vaswani2017attention}, the backbone architecture of most large language models, have emerged in several healthcare applications, primarily clinical prediction tasks for structured \glspl{ehr} \cite{nerella2024transformers}, information extraction from clinical free-text notes \cite{kalyan2022ammu} and medical imaging \cite{shamshad2023transformers}. 

Traditional \gls{dl} approaches for \gls{ehr} sequence modeling utilized convolutional neural networks \cite{nguyen2016mathtt} or \glspl{rnn} \cite{choi2016retain}. However, \glspl{rnn} struggle with long-range dependencies present in \glspl{ehr} because of vanishing gradients \cite{bengio1994learning}. Although \gls{rnn} variants like \gls{lstm} \cite{hochreiter1997long} reduce the gradient problem, its complexity scales poorly \cite{al2024rnn}. Transformers, primarily BERT (bidirectional encoder representations from Transformers) \cite{devlin2019bert}, have demonstrated their ability to capture long-range and inter-concept dependencies inherent in \glspl{ehr} efficiently through the attention mechanism \cite{vaswani2017attention}, surpassing \glspl{rnn} \cite{li2020behrt, meng2021bidirectional} and \glspl{lstm} \cite{meng2021bidirectional,  
pang2021cehr, yang2023transformehr}. 

While efficient, Transformers face quadratic training time and memory complexity as self-attention learns pairwise contextual representations between all input tokens \cite{vaswani2017attention}. The performance of Transformers also scales with the dataset size (tokens) and model size (parameters) \cite{kaplan2020scaling}. Advanced Transformers (Transformers++) like Llama \cite{touvron2023llama} incorporate architectural improvements, such as rotatory positional embeddings \cite{su2024roformer} or pre-normalization, enabling them to handle longer sequences.
Attention-free designs like Mamba \cite{gu2023mamba} achieve linear scaling by leveraging \glspl{ssm} equipped with context-aware selection mechanism and hardware-aware computation \cite{qu2024survey}. Due to its competitive performance with Transformer++, Mamba has found widespread applications in \gls{dl} fields, including language models \cite{waleffe2024empirical} or computer vision \cite{liu2024vision}. For longer \gls{ehr} sequences, Mamba has been shown to outperform advanced Transformers \cite{fallahpour2024ehrmamba, wornow2024context}.

Most previous works focus on general-purpose heterogeneous cohorts \cite{li2020behrt, wornow2024context, rasmy2021med}. In contrast, the present analysis focuses on \glspl{ehr} from a Swedish cohort with \gls{hf}, a well-defined and validated condition \cite{schaufelberger2020validity}. We argue that disease-specific models tailored around defined diagnostic conditions are more likely to translate into clinical practice given that the availability and diversity of clinical concepts remain local or lack standardization. \Gls{hf} is characterized by relatively few hospitalizations during the early stage, whereas healthcare utilization and hospitalizations increase substantially during advanced disease \cite{desai2012rehospitalization, dunlay2015care}. At the end stage, deterioration is already clinically apparent, limiting the value of prediction, whereas early prediction can still support intervention and care planning. Therefore, we focus on the initial \gls{hf} diagnosis in-hospital as a clinically meaningful early-stage milestone. Since these patient trajectories have comparatively short longitudinal histories, we tailor our study to short-context settings ($\leq$512 tokens) and tiny- to medium-sized models (2--35 M parameters), reflecting a clinically realistic model deployment scenario with moderate computational requirements. Although long-context settings capture more historical information, for this specific setting we assume that their additional computational cost may not be justified.

Our contributions summarize as follows:
\begin{itemize}
    \item \textbf{\Gls{tgds}}. We present \gls{tgds}, an \gls{ehr}-based sequence modeling framework for one-year prediction of clinical instability and mortality following the initial \gls{hf} diagnosis. Rather than introducing a new sequence architecture or representation, \gls{tgds} provides a unified formulation of compact end-to-end patient trajectories from structured in-hospital \gls{ehr}, integrating category-level clinical concept tokenization, recency-aware temporal representation through short context, and autoregressive sequence modeling with a Llama backbone trained on a \gls{ntp} objective.
    \item \textbf{Systematic investigation of sequence modeling design choices.} 
    We provide a systematic study of key design choices that are often confounded in prior \gls{ehr} sequence modeling approaches. This analysis characterizes which components contribute effectively to longitudinal \gls{hf} trajectory modeling. Specifically, we investigate the contribution of tokenization granularity, temporal representation strategies, and sequence modeling configurations, including six model architectures (e.g., BERT, Llama, Mamba), model size and context length.
    \item \textbf{\gls{tgds} improves risk prediction over established \gls{ehr} baselines.} On two one-year prediction tasks (clinical instability and mortality) in a Swedish \gls{hf} cohort ($N=42{,}820$), \gls{tgds} outperforms a well-established BERT baseline, with improved calibration. The ablation study shows that the selected design performs consistently, with a tiny Llama variant already surpassing larger BERT-based baselines, suggesting that gains arise from architecture choice rather than model scale alone. Internal hospital validation strategies further demonstrate that \gls{tgds} performs robust across independent clinical sites.
    \item \textbf{Translation into four discharge care pathways.} Joint predictions of clinical instability and mortality partition patients into four risk-based discharge pathways, providing an interpretable foundation for future clinical decision-support applications.
\end{itemize}

\section{Related work}
Transformers, particularly BERT \cite{devlin2019bert}, have been widely applied to \glspl{ehr} for predictive tasks, as demonstrated by models such as BEHRT \cite{li2020behrt}, BRLTM \cite{meng2021bidirectional}, and Med-BERT \cite{rasmy2021med}. CEHR-BERT \cite{pang2021cehr} and ExBEHRT \cite{rupp2023exbehrt} were pretrained on general patient populations and subsequently fine-tuned to predict 30-day readmission in \gls{hf} subcohorts. Similarly, Med-BERT \cite{rasmy2021med} was finetuned on patients with both diabetes and \gls{hf} for disease prediction and Antikainen et al. \cite{antikainen2023transformers} predicted 6-month mortality in cardiac patients using Transformers. By incrementally optimizing BEHRT and applying techniques similar to those in Transformer++, CORE-BEHRT \cite{odgaard2024core} identified substantial performance gains from enhanced data representations by incorporating medications and temporal encoding of patient age. CEHR-BERT \cite{pang2021cehr} enhanced the temporal representation of \gls{ehr} by introducing an artificial time token [ATT] which stores the time gap between two consecutive visits. In CEHR-BERT,  different patient representations were compared highlighting the importance of the [ATT] token, whereas Wornow et al.'s \cite{wornow2024context} findings indicate a model-dependent benefit of [ATT]. To avoid sequence truncation by the context length, ExBEHRT \cite{rupp2023exbehrt} stacked clinical concepts vertically through additional embeddings. Similarly, Hi-BEHRT \cite{li2022hi} introduced a slicing window mechanism to capture long-term dependencies between each visit segment. In G-BERT \cite{shang2019pre}, the learned ontology of medical concepts were improved through BERT embeddings combined with a graph neural network. Foresight \cite{kraljevic2024foresight} forecasts temporally-aware future clinical concepts within a patient trajectory leveraging structured and unstructured \glspl{ehr}. BEHRT was integrated with observational causal inference techniques to estimate treatment effects, such as risk ratios \cite{rao2022targeted}.

Recent \gls{ehr} modeling works have shifted towards architectures beyond BERT, often surpassing it at longer context lengths \cite{yang2023transformehr, fallahpour2024ehrmamba, wornow2024context, antikainen2023transformers}. Antikainen et al. \cite{antikainen2023transformers} demonstrated that XLNet \cite{yang2019xlnet} is a competitive alternative to BERT capturing more positive cases. TransformEHR \cite{yang2023transformehr} introduced a new pretraining objective by forecasting all diagnoses in a single visit using an encoder–decoder framework. EHRMamba \cite{fallahpour2024ehrmamba} paired Mamba with multi-task learning and outperformed Transformers at a context length of 2048. However, the study did not investigate Transformer++ or Mamba at varying context lengths. Wornow et al. \cite{wornow2024context} evaluated sequence models on various benchmark prediction tasks and \gls{ehr} properties, such as token repetition and visit irregularity, showing that Mamba is more robust at longer context lengths beyond 4k tokens but achieves a reduced discriminative performance at 1k tokens compared to Transformers++.

\section{Methods}

\subsection{Heart failure cohort}
A \gls{hf} cohort was defined consisting of patients $\geq 18$ years old at their initial \gls{hf} diagnosis in-hospital, identified by qualifying \gls{icd} codes (I110, I130, I132, any I42 or any I50 in any position). The study period for diagnostic inclusion spanned between January 1, 2015 and December 31, 2022, resulting in $N=42{,}820$ patients. To include one-year prediction outcomes (Section~\ref{ssec:tasks}) for all patients, the observation period ended December 31, 2023. The patient data was extracted from \glspl{ehr} covering all emergency in-hospital and public specialized out-patient care from six hospital sites within the Region Västra Götaland (VGR) of Sweden (total population 1.8 million). Supplementary Table~\hyperref[stab:features]{S1} shows the collected clinical concepts including demographics (age, sex, \gls{bmi}), \gls{dx}, \gls{vit}, \gls{lab}, \gls{med} and \gls{pro}. While \gls{dx} and \gls{med} were assessed by international classification systems, \gls{icd} and \gls{atc} respectively, \gls{pro} were assessed by the Swedish Classification of Care Measures (KVÅ\footnote{\url{https://www.socialstyrelsen.se/statistik-och-data/klassifikationer-och-koder/kva/}}) system. 

\begin{table*}[!h]
    \centering
    \caption{The three components of \gls{tgds}, the chosen design for each component, and the alternatives explored by ablation (Section~\ref{ssec:ablation}).}
    \label{tab:pipeline}
    \begin{tabular}{llp{4cm}p{6.5cm}}
    \toprule
     Id & Component & Chosen design (\gls{tgds}) & Alternatives explored by ablation \\
     \midrule
     1 & Tokenization & Category-level ICD-10 codes ($i=3$); 10-bin discretization of VIT, LAB, MED (\voctwo{}) & Finer/coarser bins ($b\in\{5,10\}$) and code levels ($i\in\{3,4\}$) (Section~\ref{sssec:a1}) \\
     2 & Temporal representation & Chronologically ordered events with [ATT] visit-gap tokens, right-sided cutoff bounded by a short context $C=512$ (\histcontext{}) & Truncation horizons (\histtzero{}, \histtone{}, \histtthree{}) and daily/bi-daily aggregation (\histwinone{}, \histwintwo{}) (Section~\ref{sssec:a2}) \\
     3 & Sequence model & Llama with \glsentryshort{ntp} objective, Medium size, $C=512$ & \glsentryshort{mlm} (BERT, ModernBERT), \glsentryshort{plm} (XLNet), NTP (Mamba, Mamba2); model size $\in\{$Tiny, Small, Medium$\}$; context $C\in\{128, 256, 512\}$ (Section~\ref{sssec:a3}) \\
     \bottomrule
    \end{tabular}
\end{table*}

\subsection{\Gls{tgds}: Trajectory-guided discharge stratification for heart failure} \label{subsec:pipeline}
Structured longitudinal \glspl{ehr} from real-world hospitalized \gls{hf} patients are inherently irregular multivariate time series. Patients have diverse sets of clinical concepts recorded at irregular time points and frequency, varying across patients and over the course of individual patient trajectories and ranging from sparse to dense event trajectories. These \gls{ehr} characteristics make sequence models a natural choice for learning latent patient embeddings. We develop \gls{tgds}, a methodology that reads the patient's recent in-hospital trajectory of diagnoses, vital signs, laboratories, medications, and procedures end-to-end with a compact autoregressive sequence model, to stratify the complementary one-year risks of clinical instability and mortality that drive \gls{hf} discharge care. The overarching research question is:

\begin{enumerate}[label=\textbf{RQ}]
    \item How to design a methodology that reads the recent in-hospital trajectory of a patient at their first \gls{hf} diagnosis, stratified for one-year risks of clinical instability and mortality at discharge?
\end{enumerate}

The \glspl{ehr} of a patient can be formulated as a sequence of clinical events \mbox{$e_j = (c_j, v_j, t_j, x_j)$}, where $j$ denotes the event index, $c_j$ the clinical concept, $v_j$ the visit index, $t_j$ the irregular event time point and $x_j$ the concept-specific event value. \Gls{tgds} (Table~\ref{tab:pipeline}) is defined by three explicit design choices: a category-level tokenization (Section~\ref{sssec:c1}), a recency-weighted temporal representation bounded by a short context (Section~\ref{sssec:c2}), and a Llama sequence model trained on \gls{ntp} (Section~\ref{sssec:c3}). Two separate models with the same configuration produce the predictions for clinical instability and mortality, whose joint outcome stratifies patients into four discharge care pathways (Section~\ref{ssec:tasks} and Table~\ref{tab:clinical_impact}).

\subsubsection{Tokenization (chosen: \voctwo{})} \label{sssec:c1}
\Gls{tgds} tokenizes each event value $x_j$, conditioned on its concept $c_j$, into a shared vocabulary via $\tau : (c_j, x_j) \rightarrow \hat{x}_j$. Standardized categorical codes (\gls{icd}, \gls{atc}, KVÅ) are represented at the category level ($i=3$ for \gls{icd}), and continuous numerical values (\gls{vit}, \gls{lab}, \gls{med} amounts) are discretized into $b=10$ bins. This yields the vocabulary \voctwo{} (Supplementary Table~\hyperref[stab:vocab]{S5}), which keeps the embedding space compact while preserving clinically meaningful resolution. Ablation~A1 (Section~\ref{sssec:a1}) varies $b\in\{5,10\}$ and $i\in\{3,4\}$ to verify that finer or coarser tokenization does not improve over the chosen design.

\subsubsection{Temporal representation (chosen: \histcontext{} with $C=512$)} \label{sssec:c2}
Patient-specific clinical events are sorted chronologically from the earliest to the latest timestamp to form a patient sequence $\mathcal{E} = (e_1, \dots, e_L)$, where $L$ denotes the sequence length, with [ATT] tokens inserted to mark inter-visit gaps. \Gls{tgds} bounds the sequence right-sided by a short context length $C=512$ tokens (\histcontext{}), with $C\leq L$ discarding earlier events. This recency-weighted design emphasizes events that are most predictive at discharge while remaining computationally compact. Ablation~A2 (Section~\ref{sssec:a2}) explores both shorter observation horizons (truncation at $t=0$, $t\leq 1\mathrm{y}$, $t\leq 3\mathrm{y}$) and temporal aggregation that compresses repeated continuous events into daily or bi-daily summaries (\histwinone{}, \histwintwo{}). Daily aggregation is reported as a task-specific refinement that improves mortality predictions rather than as part of the \gls{tgds} default.

\subsubsection{Sequence model (chosen: Llama with NTP)} \label{sssec:c3}
The patient sequence is represented as an embedding vector $\mathbf{E} \in \mathbb{R}^{C \times d_m}$ with $\mathbf{E}=\mathbf{E}_{\hat{x}}+\mathbf{E}_{v}+\mathbf{E}_{t}$, where $d_m$ denotes the embedding size, $\mathbf{E}_{\hat{x}}$ the same-sized embeddings of the tokenized values, $\mathbf{E}_{v}$ of the visit indices, and $\mathbf{E}_{t}$ of the timestamps. \mbox{\Gls{tgds}} uses a Llama backbone \cite{touvron2023llama} (rotary positional embeddings, RMSNorm pre-normalization, SwiGLU activations) pretrained autoregressively with an \gls{ntp} objective. The Medium configuration (hidden dimension $d_m =512$, 6 hidden layers, 8 attention heads) is the default. The autoregressive backbone with learnable weights $\theta$ maps the embedded sequence $\mathbf{E}$ to a latent patient representation $\mathbf{z}=f_\theta(\mathbf{E})$. Because clinically relevant signals in \gls{hf} are concentrated within recent activity, short-context modeling fits the task; Ablation~A3 (Section~\ref{sssec:a3}) verifies these design choices by comparing representative architectures (BERT, XLNet, ModernBERT, Llama, Mamba, Mamba2), their associated pretraining paradigms (\gls{mlm}, \gls{plm}, \gls{ntp}), model sizes, and context lengths.

\subsection{Clinical prediction tasks} \label{ssec:tasks}
A classifier with learnable weights $\phi$ predicts an outcome $\hat{y}$ for a given patient representation $\mathbf{z}$ via $\hat{y}=g_\phi(\mathbf{z})$ aligned with  clinical prediction task derived by domain experts. Patient trajectories can be defined at different entry points or endpoints, influencing clinical decision-making. For example, patient trajectories centered around a first time diagnostic index event provide different predictive insights and clinical implications than late-stage patient trajectories with progressed diseases.

\begin{figure*}[!h]
    \centering
    \includegraphics[width=\linewidth]{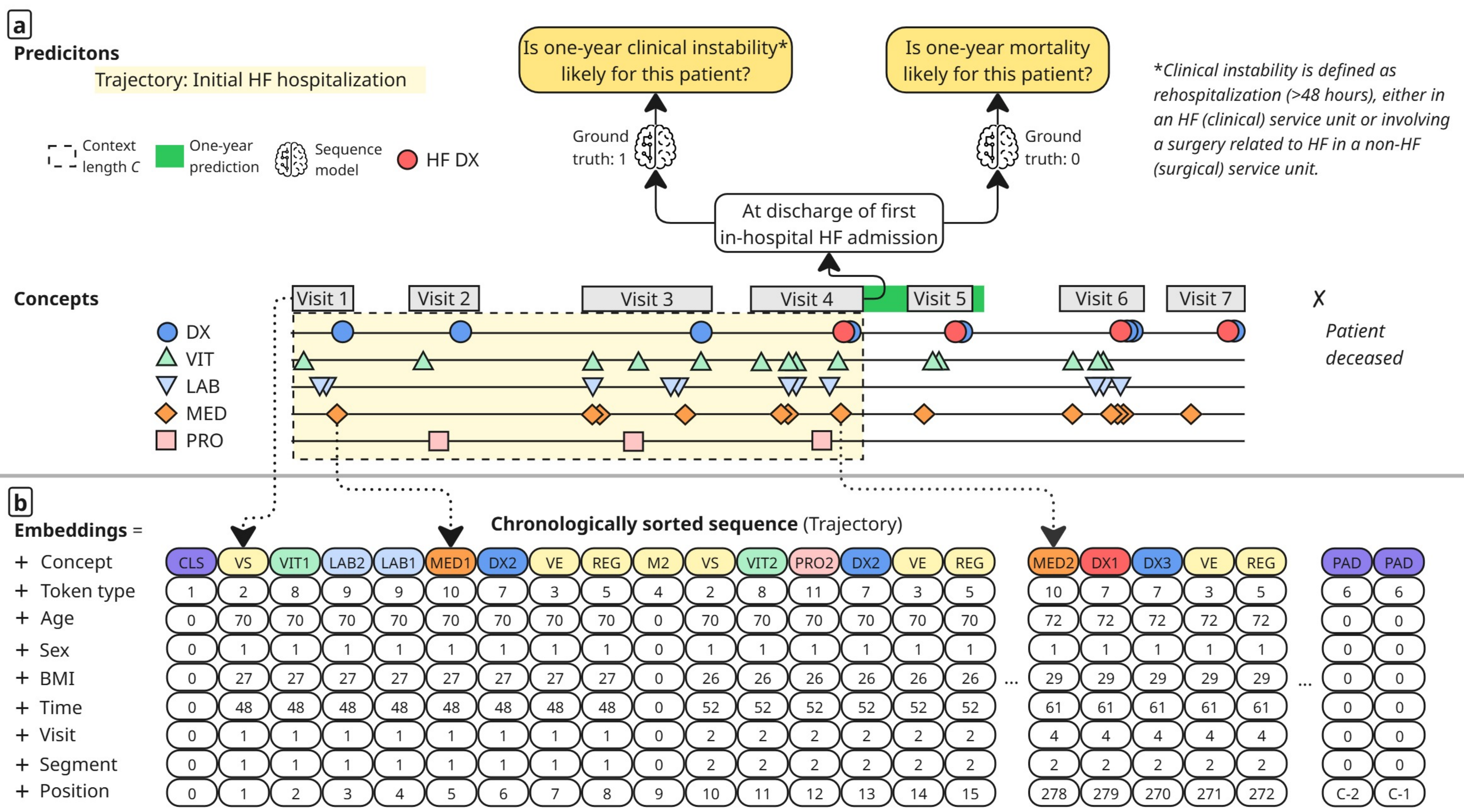}
    \caption{(\textbf{a}) Overview of the clinical prediction tasks. Given a simplified, chronologically ordered patient sequence, separate sequence models were trained to predict at discharge: one-year clinical instability and one-year mortality, both following the initial \gls{hf} diagnosis in-hospital. (\textbf{b}) Corresponding input sequence representation after the initial \gls{hf} diagnosis trajectory highlighting the aggregated embeddings of tokenized concepts, concept token types, demographics (age, sex, \gls{bmi}), time of events, visit numbers, alternating visit segments and absolute token positions.}
    \label{fig:tasks_embeddings}
\end{figure*}

Two binary classification tasks were evaluated in the \gls{hf} cohort at the initial \gls{hf} diagnosis in-hospital, which defines the patient trajectory used for prediction (Fig~\ref{fig:tasks_embeddings}a). The trajectory contains historical patient data up to and including the index hospitalization for \gls{hf}. For both tasks, the prediction window was within one year after discharge from hospital following the initial \gls{hf} diagnosis. The first task $T_1$ predicted clinical instability (prevalence $P_{T_1}=39.7\%$), a phenotype of unplanned readmission to hospital; the second task $T_2$ predicted mortality ($P_{T_2}=24.8\%$). At the initial \gls{hf} admission, the demographic profile (Supplementary Table~\hyperref[stab:cohort]{S3}) of the \gls{hf} cohort ($N=42{,}820$, 45.4\% women) was median 80 (\gls{iqr} 72, 87) years old, with a median of 153 (71, 324) recorded events per patient trajectory and 2 (1, 3) hospitalizations. 

Clinical instability was defined using a rule-based approach as rehospitalization for a period of at least 48 hours that met one of the following criteria (Supplementary Table~\hyperref[stab:crit_instability]{S2}): i) admission to a clinical service unit relevant to \gls{hf} care (e.g., cardiology or internal medicine), or ii) admission in a surgical service unit where a patient underwent a surgical procedure relevant to \gls{hf} determined by selected procedure codes (e.g., implantation of a cardiac device). The minimum duration was set to 48 hours to exclude hospitalizations which were less severe or possibly planned (e.g., routine checkups, precautionary in-hospital observation or planned interventions not caused by worsening of HF).

\subsection{Patient sequence representation}
To construct a patient sequence, hospitalizations with at least one diagnosis were chronologically ordered so that the latest event occurred right-sided. Consecutive hospitalizations less than 24 hours apart were merged into one. Hospitalizations that led to in-hospital death were excluded as not relevant for the outcomes. Out-patient encounters were excluded because they indicate low clinical severity. The events from all patient sequences were tokenized and mapped to a shared baseline vocabulary. \gls{icd} and KVÅ were registered at the third code-specific hierarchical level. Related KVÅ codes with a digit in the third position were standardized by replacing the digit with an uncommon shared character. New \gls{med} concepts were constructed by concatenation of the \gls{atc} code (level 4), the route of administration (e.g., intravenous or oral) and the amount of medication. All \gls{vit}, \gls{lab} and \gls{med} concepts were independently discretized into 10 quantile bins. No unit normalization was applied, as all continuous measurements were recorded using standardized units, while \gls{med} were encoded as administration route-specific concepts, ensuring consistent units. Missing \gls{bmi} values were median imputed, as \gls{bmi} was part of the demographic embeddings and not a concept of the patient trajectory.

Fig.~\ref{fig:tasks_embeddings}b shows the embeddings of an example patient sequence. Each time-irregular visit contains a varying number and types of clinical concepts. The sequence length is determined by the context length $C$, hence, events that have occurred earlier might be discarded. The embeddings are an aggregation of the same-sized individual embeddings of the concepts and token types to contextualize the events, demographic triplet (age, sex, body mass index), passed time in weeks given a reference date, visit number of the respective hospitalizations, visit segments to differentiate between consecutive hospitalizations and absolute position within the sequence. Following previous works \cite{pang2021cehr, fallahpour2024ehrmamba}, the tokenization includes special tokens like [CLS] and [PAD] to mark sequence boundaries, visit-specific tokens [VS] and [VE] to signal start and end of hospitalizations, and the artificial time token [ATT] to indicate temporal gaps between consecutive hospitalizations (e.g., [M2] for a two-month gap). The register token [REG] can be interpreted as a storage for global information after each encounter \cite{darcet2023vision}.

\subsection{Implementation}
This work investigated six sequence models across three architecture classes: (i) Transformers (BERT \cite{devlin2019bert}, XLNet \cite{yang2019xlnet}), (ii) Transformers++ (ModernBERT \cite{warner2024smarter} abbreviated as MBERT, Llama \cite{touvron2023llama}), and (iii) Mambas (Mamba \cite{gu2023mamba}, Mamba2 \cite{dao2024transformers}). Their associated learning objectives are: \gls{mlm} (BERT, MBERT), \gls{plm} (XLNet), \gls{ntp} (Llama, Mamba, Mamba2). To compare with traditional machine learning, the commonly used \gls{xgb} \cite{chen2016xgboost} was trained on the frequency of clinical concepts within a patient sequence \cite{pang2021cehr, fallahpour2024ehrmamba, rupp2023exbehrt}. Supplementary Section~\hyperref[ssec:sequence_models]{S2.1} 
describes the architectural variations across the sequence models while Supplementary Section~\hyperref[ssec:model_theory]{S2.2} provides a theoretical comparison between Transformer and Mamba. The implementation details, including hardware and software versions, data split, and training procedure, are outlined in Supplementary Section~\hyperref[ssec:implementation]{S2.3}.

The prevalence of outcomes in the \gls{hf} cohort (Supplementary Table~\hyperref[stab:cohort]{S3}) is low, e.g., the one-year mortality following the initial \gls{hf} diagnosis is $24.8\%$. Therefore, the \gls{auprc} was reported as primary metric for discriminative evaluation instead of the \gls{auroc} to better capture the model's ability to identify true positives in the presence of class imbalance. The calibration capability was assessed using the Brier score which measures the mean squared error between predicted probabilities and actual outcomes. All metrics were reported with $95\%$ \glspl{ci}, computed via bootstrapping (1000 iterations).

\section{Experiments and results} \label{sec:experiments_results}
We first report the main result addressing the overarching \textbf{\gls{rq}} (\gls{tgds} vs.\  baselines), followed by the ablation study (\textbf{RQ1}--\textbf{RQ4}) that perturbs one component of \gls{tgds} at a time. Unless otherwise specified, all models were trained at the \gls{tgds} defaults: vocabulary \voctwo{}, cutoff temporal representation \histcontext{}, context length $C=512$, and model size $S=\mathrm{Medium}$. As baselines, we used the classical \gls{xgb} on bag-of-concept frequencies \cite{pang2021cehr, fallahpour2024ehrmamba, rupp2023exbehrt} and a BERT backbone motivated by the strong performance of Transformer-based \gls{ehr} sequence models such as BEHRT, Med-BERT, and CEHR-BERT. We further included ModernBERT (MBERT), an improved BERT architecture, and XLNet, a common BERT alternative, as baselines.

\subsection{Main results}
Table~\ref{tab:key_findings} reports \gls{auprc} for \gls{tgds} (Llama, \gls{ntp}, \histcontext{}, Medium) and the baselines; Supplementary Table~\hyperref[stab:key_findings_add]{S10} repeats the comparison for \gls{auroc} and Brier score. \Gls{tgds} achieves \gls{auprc} of $0.555$ ($T_1$) and $0.574$ ($T_2$), improving over the trivial \gls{xgb} baseline by $+0.036$ and $+0.049$ \gls{auprc}, over MBERT, the strongest advanced baseline, by $+0.021$ and $+0.017$ \gls{auprc}, with confidence intervals shifting upward across both tasks (Table~\ref{tab:key_findings}). Calibration (Brier score) is similarly improved (Supplementary Table~\hyperref[stab:key_findings_add]{S10}). A task-specific refinement of \gls{tgds} using daily aggregation (\histwinone{}) further improves the mortality task to \gls{auprc} $0.582$ ($T_2$); this refinement is part of Ablation~A2 rather than the method default. Tiny variants of Llama and Mambas already surpass the Medium BERT/XLNet/MBERT baselines on both tasks while requiring substantially fewer model parameters and \gls{gmacs}, so the gain attributable to \gls{tgds} cannot be reduced to model size alone.

\begin{table*}[!h]
\centering
\caption{Main result: \gls{tgds} (Llama, \gls{ntp}, \histcontext{}, Medium) vs.\ baselines (\gls{xgb} as trivial baseline; BERT, XLNet, MBERT as advanced baselines), and additional \gls{tgds} variants (Tiny size, task-specific daily aggregation for mortality). All entries use $C=512$. \textbf{Best} \gls{auprc} (95\% \gls{ci}) per task in bold, \underline{second-best} underlined. Para.: Model parameters in M.}
\label{tab:key_findings}
\setlength{\tabcolsep}{4pt}
\begin{tabular}{@{}llllrrll@{}}
\toprule
Model & Objective & Temporal & Model size & Para. &  \multicolumn{1}{l}{GMACs} & \multicolumn{1}{c}{$T_1$} & \multicolumn{1}{c}{$T_2$} \\ \midrule
\multicolumn{8}{@{}l}{\textit{Baselines}} \\
 XGBoost & ---& Cutoff & --- & --- & --- &  0.519 \ci{(0.497--0.540)} & 0.525 \ci{(0.500--0.550)} \\
 BERT & MLM & Cutoff & Medium & 24.1 & 96.2 & 0.535 \ci{(0.516--0.556)} & 0.540 \ci{(0.516--0.564)} \\
 XLNet & PLM & Cutoff & Medium & 28.4 & 62.1 & 0.529 \ci{(0.508--0.549)} & 0.558 \ci{(0.533--0.584)} \\
MBERT & MLM & Cutoff & Medium & 32.2 & 114.2 & 0.534 \ci{(0.513--0.556)} & 0.557 \ci{(0.531--0.583)} \\
\multicolumn{8}{@{}l}{\textit{\gls{tgds} (proposed)}} \\
\textbf{TGDS-HF (Llama)} & NTP & Cutoff & Medium & 34.4 & 113.3 & \textbf{0.555} \ci{(0.535--0.575)} & \underline{0.574} \ci{(0.550--0.599)} \\
\multicolumn{8}{@{}l}{\textit{Alternative NTP backbones (ablation~A3)}} \\
Mamba & NTP & Cutoff & Medium & 25.5 & 81.7 & 0.543 \ci{(0.523--0.565)} & 0.559 \ci{(0.534--0.585)} \\
 Mamba2 & NTP & Cutoff & Medium & 26.6 & 79.9 & 0.542 \ci{(0.521--0.562)} & 0.572 \ci{(0.547--0.596)} \\
\multicolumn{8}{@{}l}{\textit{Task-specific refinement: daily aggregation for the mortality task ($T_2$)}} \\
 \textbf{TGDS-HF (Llama, agg.)} & NTP & Agg. (1d) & Medium & 34.4 & 113.3 & 0.549 \ci{(0.529--0.570)} & \textbf{0.582} \ci{(0.558--0.608)} \\
 Mamba (agg.) & NTP & Agg. (1d) & Medium & 25.5 & 81.7 & 0.543 \ci{(0.523--0.564)} & 0.571 \ci{(0.547--0.596)} \\
\multicolumn{8}{@{}l}{\textit{Tiny variants of \gls{tgds} already surpass Medium baselines}} \\
 \textbf{TGDS-HF (Llama, Tiny)} & NTP & Cutoff & Tiny & 12.9 & 25.1 & \underline{0.550} \ci{(0.529--0.570)} & 0.566 \ci{(0.540--0.592)} \\
 Mamba (Tiny) & NTP & Cutoff & Tiny & 3.3 & 3.8 & 0.546 \ci{(0.526--0.566)} & 0.561 \ci{(0.534--0.587)} \\ \bottomrule
\end{tabular}
\end{table*}

\subsection{Ablation study} \label{ssec:ablation}
The ablation study perturbs one component of \gls{tgds} at a time, holding the rest at the default (Llama, \gls{ntp}, \voctwo{}, \histcontext{}, $C=512$, Medium). The four sub-questions guiding the study are:

\begin{enumerate}[label=\textbf{RQ\arabic*}, start=1]
    \item How effective is the chosen tokenization \voctwo{} compared to finer or coarser alternatives across $b\in\{5,10\}$ and $i\in\{3,4\}$?
    \item How effective is the chosen temporal representation \histcontext{} compared to truncation and aggregation alternatives?
    \item How does the sequence-model configuration, including backbone, model size, and context length, shape the effectiveness on short-context patient trajectories?
    \item How gracefully does \gls{tgds} degrade under reduced clinical concept availability and limited training data?
\end{enumerate}

Visualizations of \gls{auroc} and Brier score along with all numerical metrics are reported in Supplementary Section~\hyperref[ssec:results]{S3}.

\subsubsection{A1: Tokenization (\textbf{RQ1})} \label{sssec:a1}
The granularity of the shared event vocabulary is perturbed in two ways: the discretization resolution of \gls{vit}, \gls{lab} and \gls{med} varies with $b\in\{5,10\}$, and the hierarchical \gls{icd} code level varies with $i\in\{3,4\}$. The category-level encoding ($i=3$) keeps the embedding space compact while preserving clinically meaningful resolution \cite{rupp2023exbehrt, odgaard2024core}; the diagnosis level ($i=4$, e.g.\ I509) provides more precise assessments by potentially identifying detailed phenotypes within the HF cohort, at the cost of a larger vocabulary. Supplementary Table~\hyperref[stab:vocab]{S5} reports the resulting vocabulary sizes.

\begin{figure}[!h]
    \centering
    \includegraphics[width=\linewidth]{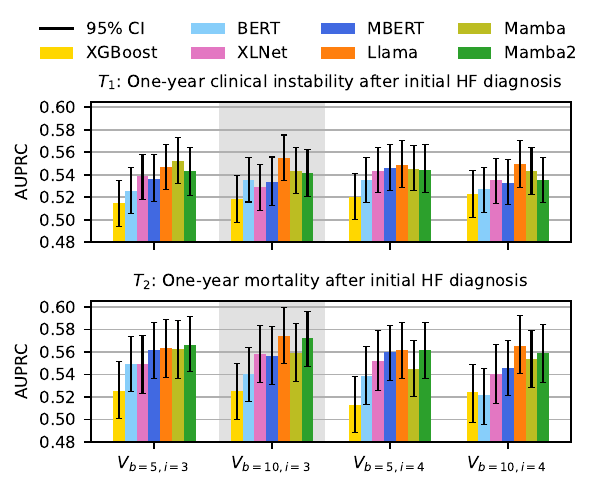}
    \caption{Ablation of the vocabulary $V$ with increasing resolution from lowest (left) to highest (right). All modifications are evaluated by bootstrapped \gls{auprc} ($\uparrow$) for $\mathrm{Medium}$-sized sequence models and $C=512$. Gray background highlights shared setup across all ablations.}
    \label{fig:ex1}
\end{figure}

Fig.~\ref{fig:ex1} reports \gls{auprc} for the four combinations, ordered by increasing unique-token count. For Llama (the \gls{tgds} backbone), the chosen \voctwo{} is at or above the best for $T_1$ across all four vocabularies and competitive on $T_2$; the finest vocabulary \vocfour{} does not improve over \voctwo{}, consistent with the conjecture that finer encodings add noise rather than predictive power. Mamba2 shows a similar pattern, while BERT, XLNet, and MBERT show no generalizable trend, with XLNet being the most sensitive to the tokenization choice. Supplementary Fig.~\hyperref[sfig:ex1]{S1} confirms the same ordering on \gls{auroc} and Brier score. The chosen \voctwo{} in \gls{tgds} is therefore supported by the ablation.

\subsubsection{A2: Temporal representation (\textbf{RQ2})} \label{sssec:a2}
The temporal representation of the patient history $H$ is perturbed by truncation and aggregation before tokenization, with vocabulary \voctwo{} and $C=512$ held fixed. The chosen cutoff \histcontext{} is the right-sided patient history bounded by $C$, where prior events are discarded. Truncation (\histtzero{}, \histtone{}, \histtthree{}) limits the look-back to the most recent hospitalization or additionally includes $\{1, 3\}$ years of prior records, motivated by if a systematic design can outperform \histcontext{}. A short truncation horizon prioritizes recent events, whereas a long one can capture temporal dependencies. Aggregation (\histwinone{}, \histwintwo{}) compresses repeated continuous events (\gls{vit}, \gls{lab}, \gls{med}) by averaging within $w\in\{1, 2\}$ hospitalization days; a longer $w$ reduces the number of tokens but extends the prior history covered. Although heuristic, the  choices for $w$ were guided by clinical specialists to represent adjacent temporal resolutions for in-hospital care, balancing the preservation of clinically meaningful trajectories against excessive granularity. With $w=1$ day-to-day changes in patient status are preserved, whereas $w=2$ trades temporal resolutions against additional historical coverage. Larger $w$ would compress ongoing events substantially, potentially missing clinically relevant changes. Supplementary Table~\hyperref[stab:hist]{S6} summarizes the temporal perturbations.

\begin{figure}[!h]
    \centering
    \includegraphics[width=\linewidth]{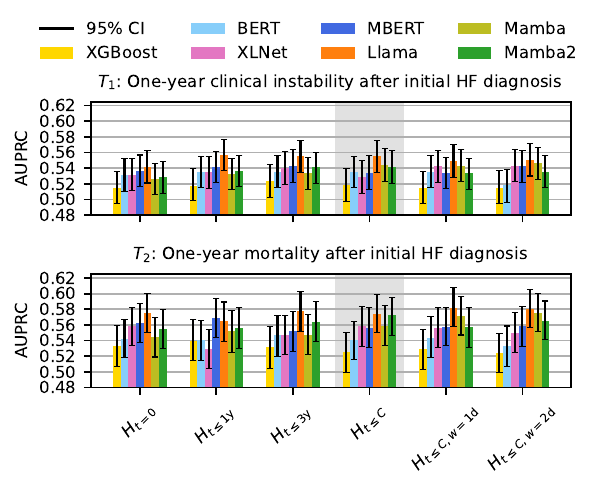}
    \caption{Ablation of the patient history $H$ with increasing historical records from the latest available hospitalization (left) to a prolonged context (right). All modifications are evaluated by bootstrapped \gls{auprc} ($\uparrow$) using $\mathrm{Medium}$-sized sequence models and $C=512$. Gray background highlights shared setup across all ablations.}
    \label{fig:ex3}
\end{figure}

Fig.~\ref{fig:ex3} reports \gls{auprc} for these temporal modifications. For Llama (the \gls{tgds} backbone), the chosen \histcontext{} is competitive across both tasks: $T_1$ instability is best modeled by the cutoff, where Llama with \histcontext{} matches or exceeds all truncations and aggregations; for the $T_2$ mortality task, daily aggregation (\histwinone{}) provides an additional improvement, which we therefore report as a task-specific refinement of \gls{tgds} rather than as part of the default. Truncations show model-specific gains in BERT (\histtthree{}) and MBERT (\histtone{}) but do not match \histcontext{} in Llama for $T_1$. Mamba2 prefers cutoff, while Mamba benefits from daily aggregation in mortality. The ablation thus supports the cutoff choice in \gls{tgds} as a single defensible default, with daily aggregation acknowledged as the recommended refinement when mortality is the primary target.

\subsubsection{A3: Sequence model, size, and context (\textbf{RQ3})} \label{sssec:a3}
Three axes are perturbed jointly: architecture (BERT, XLNet, ModernBERT, Llama, Mamba, Mamba2) together with its associated learning objective (\gls{mlm}, \gls{plm}, \gls{ntp}), context length $C\in\{128, 256, 512\}$, and architecture size $S\in\{\mathrm{Tiny}, \mathrm{Small}, \mathrm{Medium}\}$ (abbreviated $\mathrm{T}$, $\mathrm{S}$, $\mathrm{M}$). We focus on short contexts ($C\leq 512$) as the realistic clinical operating regime. Different architectures exhibit distinct characteristics for $C$, reflecting their original design goals, from shorter-context models (e.g., BERT) to long-range sequence modeling (e.g., Mamba). The vocabulary \voctwo{} is held fixed. Architecture size $S$ is controlled by the hidden dimension $d_m$, feed-forward dimension $d_f$, number of hidden layers $n_m$, and number of attention heads $n_h$ (Transformers) or Mamba blocks $n_b$ (Mambas), with hyperparameters in Supplementary Table~\hyperref[stab:modelcfg]{S8} as well as model parameter counts and \gls{gmacs} in Supplementary Table~\hyperref[stab:modelsize]{S9}. Supplementary Table~\hyperref[stab:context]{S7} shows how $C$ shapes the input sequence. To assess practical deployability, profiling results for all model configurations are reported in Supplementary Table~\hyperref[stab:train_time]{S26}, which summarizes total training time, and Supplementary Fig.~\hyperref[sfig:inference]{S6}, which presents inference runtime and peak GPU memory consumption.

\begin{figure}[!h]
    \centering
    \includegraphics[width=\linewidth]{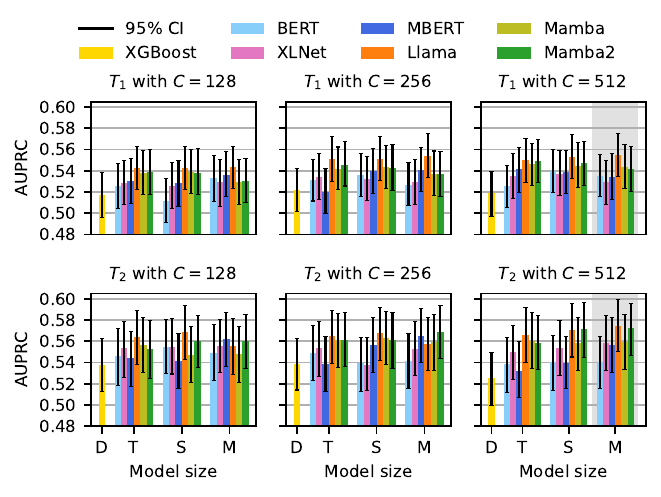}
    \caption{Ablation of the model configuration through the context length $C$ and model size. Within each $C$ sequence modes are sorted by $\mathrm{\underline{T}iny}$, $\mathrm{\underline{S}mall}$, and $\mathrm{\underline{M}edium}$ configuration,
    and compared to \gls{xgb}'s  $\mathrm{\underline{D}efault}$ configuration. All modifications are evaluated by bootstrapped \gls{auprc} ($\uparrow$). Gray background highlights shared setup across all ablations.}
    \label{fig:ex2}
\end{figure}

Fig.~\ref{fig:ex2} reports \gls{auprc} for all combinations. The \gls{tgds} default configuration (Llama, $C=512$, Medium) achieves the best \gls{auprc} on both $T_1$ and $T_2$, further validating the Llama backbone choice. The ablation also shows that the gain does not come from model size: $\mathrm{T}$-Llama at $C=128$ already surpasses the best $\mathrm{M}$-BERT/XLNet/MBERT baselines at any $C$ (except $\mathrm{M}$-MBERT at $C=256$ in $T_2$) across both tasks, and $\mathrm{T}$-Llama outperforms  any Mambas for $C\leq 256$ (except $\mathrm{M}$-Mamba2 at $C=256$ in $T_2$). Extending $C$ helps more than increasing $S$ at the Tiny scale ($\mathrm{T}$-Llama/$\mathrm{T}$-Mambas at $C=256$ beat their $C=128$ counterparts), and $\mathrm{S}$-Llama/$\mathrm{S}$-Mambas at $C=256$ already approach the Medium variants, suggesting that performance gains stem primarily from data quantity (extended $C$) as opposed to representational capacity (increased $S$). For deployment, the Medium default is a conservative choice and that Small variants are a viable lightweight version of \gls{tgds}. MBERT shows parameter inefficiency in smaller configurations, with only $\mathrm{M}$-MBERT consistently beating BERT and XLNet.

\subsubsection{A4: Robustness to data availability (\textbf{RQ4})} \label{sssec:a4}
Two robustness studies assess how gracefully \gls{tgds} degrades under reduced data availability. Both are run at the \gls{tgds} defaults ($C=512$, Medium, \voctwo{}); for the concept augmentation, concept-specific vocabularies are extracted from \voctwo{}.

\begin{figure}[!h]
    \centering
    \includegraphics[width=\linewidth]{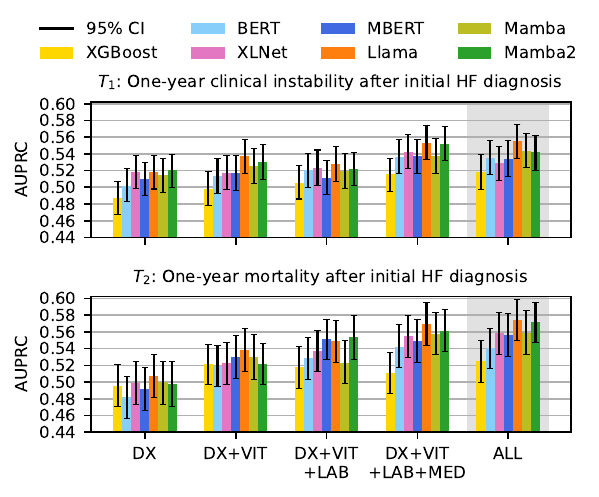}
    \caption{Incremental augmentation of \gls{dx} with  clinical concepts until all, including \gls{pro}, are added. All modifications are evaluated by bootstrapped \gls{auprc} ($\uparrow$) using $\mathrm{Medium}$-sized sequence models and $C=512$. Gray background highlights shared setup across all ablations.}
    \label{fig:ex4}
\end{figure}

\textit{Concept augmentation.} We vary the available clinical concept groups by incrementally augmenting \gls{dx} with \gls{vit}, \gls{lab}, \gls{med}, and \gls{pro} (Fig.~\ref{fig:ex4}). The order \gls{dx}, \gls{vit}, \gls{lab}, \gls{med}, \gls{pro} reflects the accessibility and resource cost of \gls{ehr} concepts: \gls{dx} and \gls{vit} are routinely recorded at low cost, whereas \gls{lab}, \gls{med} and \gls{pro} require integration from external data sources. Concepts were augmented only within the pre-defined admissions forming the patient sequence, without introducing additional admissions or extending the temporal information beyond $C$. Llama (the \gls{tgds} backbone) holds the best \gls{auprc} across nearly all task and concept combinations. Adding all five concept groups benefits all models on $T_2$, while \gls{dx}+\gls{vit}+\gls{lab}+\gls{med} is competitive on $T_1$, suggesting that \gls{pro} contributes primarily noise on the instability task. \Gls{tgds} therefore degrades gracefully as concept availability is reduced; in deployments with fewer recorded concept groups, the dominant signal is preserved.

\begin{figure}[!t]
    \centering
    \includegraphics[width=\linewidth]{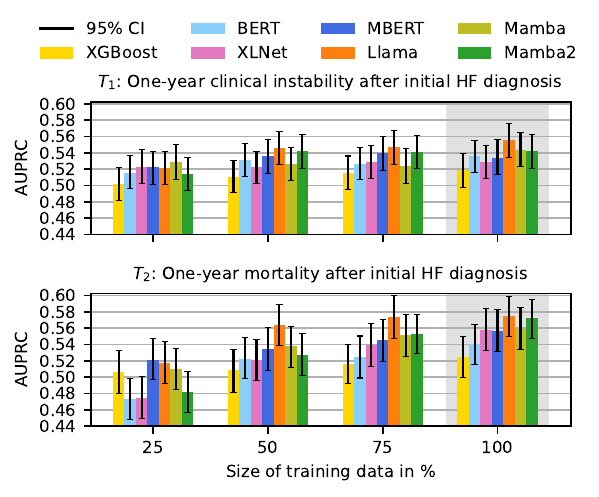}
    \caption{Ablation of different training sizes evaluated by bootstrapped \gls{auprc} ($\uparrow$) using $\mathrm{Medium}$-sized sequence models and $C=512$. Gray background highlights shared setup across all ablations.}
    \label{fig:ex5}
\end{figure}

\textit{Training-set size.} We vary the training-set fraction across $\{25\%, 50\%, 75\%, 100\%\}$ of the development set (Fig.~\ref{fig:ex5}). Most sequence models at $50\%$ training data already outperform XGBoost trained on $100\%$. Llama at $75\%$ data surpasses other models trained on $100\%$ on both tasks, and Llama converges between $75\%$ and $100\%$ on $T_2$, indicating that \gls{tgds} reaches its capacity with substantially less data than the baselines require. Mamba2 plateaus from $50\%$ on $T_1$, while Mamba benefits more from the full training set. \Gls{tgds} therefore degrades gracefully under reduced training data and remains the strongest predictor in the data-limited regime.

\subsection{Internal hospital validation}

\subsubsection{Validation strategies}
To validate the robustness of our methodology, we studied \gls{tgds} in two complementary internal hospital validation strategies against other design choices as the \gls{hf} cohort consists of \glspl{ehr} from six regional hospitals. Public datasets often differ substantially from the characteristics of the \gls{hf} cohort, such as demographics, patient trajectories, or treatment strategies, and underlying \gls{ehr} infrastructures. Consequently, internal hospital validation provides a more clinically relevant assessment. First, in \gls{shv} models were trained and evaluated using data from a single hospital to assess the predictive power in individual clinical sites. Second, in \gls{chv}
models were trained on four hospitals (H1--H4) and evaluated on two held-out hospitals (H5 and H6) to analyze the robustness across independent clinical sites. Patients were assigned to hospitals based on the location of their first \gls{hf} diagnosis in-hospital. Sites H5 and H6 were chosen as held-out sites to represent a small and medium-sized hospital. Supplementary Table~\hyperref[stab:val_hosp_stats]{S27} presents the descriptive statistics stratified by hospital for both strategies. 

For \gls{shv}, the architecture of \gls{tgds} was varied to retrospectively investigate the architecture choice ($C=512$, $S=\mathrm{Medium}$, \voctwo{}). For \gls{chv}, a heuristic subset of design choices was additionally evaluated due to computational constraints. The selection was guided by the principles of (i) covering all sequence architectures and (ii) evaluating each major ablation in at least two independent configurations. Within these constraints, the specific architecture-ablation pairings where chosen pragmatically rather than being optimized. To reduce selection bias, the subset was not restricted to the best-performing setups identified in the full-cohort experiments.

\subsubsection{Validation performance}
In \gls{shv}, \Gls{tgds} (Llama) achieves the best \gls{auprc} on average for the clinical instability prediction and is among the best sequence models behind \gls{xgb} for the mortality prediction (Supplementary Table~\hyperref[stab:eval_shv_auprc]{S28}). The \gls{shv} highlights that \gls{xgb} tends to outperform sequence models in hospitals with few patients (e.g, H1 or H5), consistent with sequence models benefiting from larger training cohorts. 

\begin{figure*}[!t]
    \centering
    \includegraphics[width=\textwidth,scale=0.75]{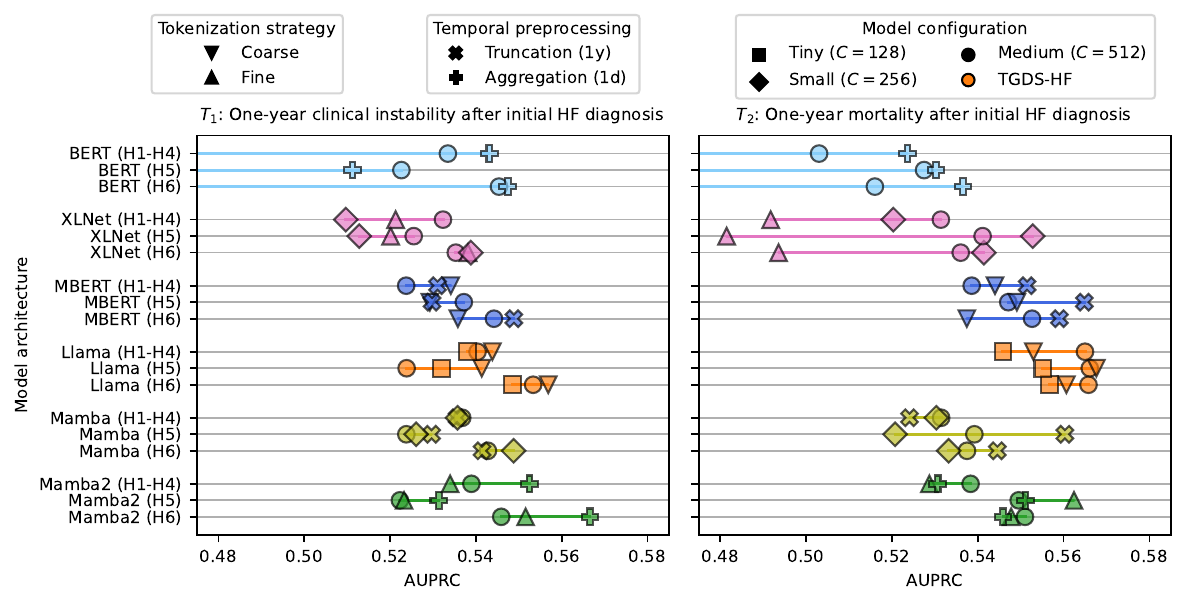}
    \caption{The \gls{auprc} ($\uparrow$) performance of different model configurations in the cross-hospital validation strategy, where only hospitals H1--H4 are part of the training set and hospitals H5 and H6 are held-out test sets. Unless stated otherwise, all models were trained with size $\mathrm{Medium}$, $C=512$ and \voctwo{}.}
    \label{fig:val_hosp_cross}
\end{figure*}

Fig.~\ref{fig:val_hosp_cross} shows the \gls{auprc} values using the \gls{chv} strategy for selected setups, with detailed comparison provided in Supplementary Table~\hyperref[stab:eval_chv_auprc]{S29}. Overall, all configurations demonstrated consistent performance across internal (H1--H4) and held-out sites (H5 and H6), suggesting robustness across clinical sites. Specifically, \gls{tgds} and its variants remain among the top-performing configurations for both tasks across the three hospital test sets. Although BERT or Mamba2 with one-day aggregation perform better than \gls{tgds} on $T_1$ for internal sites, they do not maintain this result in $T_2$, further highlighting the clinical value of \gls{tgds}. BERT and XLNet setups perform worst with  high \gls{auprc} variability, while MBERT ones show task-dependent differences. Mamba configurations achieve consistent moderate scores regardless of design compared to the larger variability observed in Mamba2.

\section{Discussion}

\subsection{Clinical implications}
\Gls{hf} is characterized by impaired cardiac function and reduced quality of life \cite{groenewegen2020epidemiology}, with prevalence increasing steeply with age and frequently coexisting with, and causally interacting with, other cardiometabolic disorders and comorbidities \cite{van2014co, azad2014management}, rather than originating from single clinical event. Clinical instability and mortality represent two major adverse outcomes after the initial \gls{hf} hospitalization. Their early identification may enable timely interventions, improve patient’s quality of life and optimize the allocation of healthcare resources. \Gls{tgds} leverages longitudinal \gls{ehr} sequences to capture temporal disease progression for accurate predictions, which provides a more comprehensive patient representation than static features. Furthermore, routinely collected EHR data can be utilized with low infrastructure requirements compared to other modalities, such as medical imaging or clinical text.

Although related, clinical instability and mortality capture complementary aspects of \gls{hf} progression. Clinical instability reflects the risk of recurrent decompensation and healthcare utilization, whereas mortality reflects the progression toward advanced disease. Combined, these predictions provide a comprehensive characterization of patient risk and enable differentiated care pathways. \Gls{tgds} is intended to be a decision-support tool that complements and not replaces clinical judgment. The predicted risks may support individualized follow-up planning, prioritization of healthcare resources, and the selection of appropriate discharge care pathways (Table~\ref{tab:clinical_impact}).

\begin{table*}[!t]
\centering
\caption{Clinical decision-making pathways based on the respective predicted one-year outcomes after the initial \gls{hf} diagnosis in-hospital.}
\label{tab:clinical_impact}
\begin{tabular}{@{}p{2.25cm}p{1.25cm}p{2.5cm}p{8.5cm}@{}}
\toprule
Clinical instability prediction & Mortality prediction & Decision of care pathway & Implication of care plan management \\ \midrule
Stable & Survived & Primary care & The patient undergoes standard treatment in primary care. Emphasis is placed on early detection of potential deterioration through vital sign assessment and management of chronic conditions. \\
Unstable & Survived & Specialized care & The patient is referred to specialized care for targeted intervention and close monitoring. Emphasis is placed on mitigating risk of complications caused by potential acute deterioration through advanced diagnostics and timely therapeutic adjustments. \\
Stable & Deceased & Regular home care & Although the patient is predicted to be clinically stable, the mortality prediction suggests underlying disease progression. Enrollment in a regular home care program focuses on symptom control, continuity of care, and quality of life improvement.  \\
Unstable & Deceased & Intensive home care & The patient is identified as high risk. Enrollment in an intensive home care program focuses on symptom relief, prevention of avoidable crises, and quality of life improvement. \\
\bottomrule
\end{tabular}
\end{table*}

Unlike most studies that trained models on general-purpose cohorts \cite{li2020behrt, odgaard2024core, fallahpour2024ehrmamba, wornow2024context} or subsequently fine-tuned for disease-specific subcohorts \cite{meng2021bidirectional, rasmy2021med, rupp2023exbehrt}, \gls{tgds} is trained directly on a disease-specific cohort and captures meaningful relationships for the two clinically relevant prediction tasks. From a real-world perspective where access to general-purpose cohorts can be limited, disease-specific modeling is justified as \gls{hf}-specific trajectories, clinical concepts and comorbidity patterns may be underrepresented in general-purpose cohorts.

\subsection{Key findings}
The overarching \gls{rq} is answered affirmatively: \gls{tgds} outperforms both the trivial \gls{xgb} baseline and the advanced baselines (BERT, XLNet, ModernBERT) across both one-year tasks, with overlapping or improved calibration. Each of its three design choices is independently supported by the corresponding ablation. The category-level tokenization \voctwo{} (Ablation~A1) matches or exceeds finer or coarser vocabularies. The recency-weighted cutoff \histcontext{} (Ablation~A2) is competitive across tasks, with daily aggregation reported as a task-specific refinement for mortality. The \gls{ntp}-Llama backbone at $C=512$ and Medium size (Ablation~A3) outperforms \gls{mlm}- and \gls{plm}-based architectures on both $T_1$ and $T_2$; the Tiny variant already surpasses larger \gls{mlm}/\gls{plm} baselines, so the gain is attributable to the design rather than to model size. Finally, \gls{tgds} degrades gracefully under reduced concept availability and training data (Ablation~A4).

\subsection{Why each \gls{tgds} design choice holds up}
\textit{Tokenization (A1, Fig.~\ref{fig:ex1}).} The chosen \voctwo{} is supported across both tasks: the finest vocabulary \vocfour{} performs worst, consistent with prior reports that more granular diagnosis and medication codes contribute only minor gains \cite{odgaard2024core} and indicating that excessive resolution introduces noise rather than predictive power. BERT and MBERT show poor generalization across tokenizations, whereas XLNet is most sensitive to the choice.

\textit{Temporal representation (A2, Fig.~\ref{fig:ex3}).} The chosen \histcontext{} is competitive across both tasks. For the $T_2$ mortality task, daily aggregation (\histwinone{}) compresses repeated continuous events into semantically meaningful summaries and provides an additional gain, suggesting temporal dependencies are more important than event quantity. \Gls{tgds} reports this as a task-specific refinement: clinicians targeting mortality stratification should adopt \histwinone{}, while \histcontext{} remains the unified default.

\textit{Sequence model, size, and context (A3, Fig.~\ref{fig:ex2}).} The chosen Llama backbone is supported by the architecture ablation. Tiny Llama and Mambas consistently outperform the best BERT/XLNet/MBERT configurations at matched $C$, so the gain is driven by the contextual learning capability of the \gls{ntp} objective, not by model size. The autoregressive token generation may better capture the progression of patient trajectories, as sequence coherence learned through \gls{ntp} can preserve clinically relevant risk signals. Among the \gls{ntp} models, advanced attention mechanisms appear to learn representations more efficiently than state-space formulations in short-context settings ($\leq$ 512 tokens), consistent with prior observation that Llama performs better than Mamba in similar shorter $C$ ranges \cite{wornow2024context}. The Small variant at $C=256$ approaches the Medium default, offering a lightweight deployment option for \gls{tgds}. Combined, these findings support that Llama is effective for short-context \gls{ehr} settings.

\textit{Robustness (A4, Figs.~\ref{fig:ex4}--\ref{fig:ex5}).} \Gls{tgds} degrades gracefully along both data axes: concept diversity dominates training-set size (Llama at $75\%$ with all concepts beats $100\%$ with fewer concepts), and adding \gls{pro} is not always beneficial, possibly because the constructed standardization of KVÅ codes fails to account for hierarchical levels and could be improved with semantic similarity based on code descriptions. \Gls{tgds} maintains its lead with $25\%$ less data than the advanced baselines need to match their full-data performance. Further, these findings highlight the importance of diverse clinical concepts and appropriate architectural choice rather than a large dataset.

\subsection{Robustness across hospital settings}
\Gls{tgds} demonstrated robustness across both \gls{shv} (Supplementary Table~\hyperref[stab:eval_shv_auprc]{S28}) and \gls{chv} (Fig.~\ref{fig:val_hosp_cross}) strategies, aligning with the main findings observed in the full-cohort study (Table~\ref{tab:key_findings}). The \gls{chv} analysis further suggests that differences between configurations are primarily architecture-dependent and secondarily configuration-specific, with greater variability found in architectures trained on objectives other than \gls{ntp}. Consequently, selecting a sequence architecture appropriate for the clinical cohort and prediction tasks appears more important than extensive optimization of individual design choices.

\subsection{Limitations and future work}
There are several limitations in our work. Firstly, \gls{tgds} was developed and evaluated on a Swedish \gls{hf} cohort for a specific clinical milestone with internal hospital but no external validation; although \gls{hf} represents a heterogeneous population, the absolute performance may not transfer to other healthcare systems. Secondly, laboratories and medications were selectively curated for \gls{hf}, which potentially limits the model from learning out-of-domain knowledge. Alternative patient embeddings, such as the pre-concatenation of demographic information into the sequence instead of separate embeddings, were not explored. Thirdly, \gls{tgds} establishes a predictive foundation for future decision-support applications; however, its clinical utility and implementation into clinical workflows have not yet been evaluated. Prospective studies involving clinicians and real-world workflow integration are required to determine whether \gls{tgds} predictions can improve clinical decision-making. Future work should fuse \gls{tgds} token sequences with multimodal \gls{ehr} signals, such as cardiac images or clinical text notes, to explore additional performance benefits. The inclusion of information for medical disciplines and out-of-hospital visits could prospectively enrich the trajectory representation used in \gls{tgds}.

\section{Conclusion}
We presented \gls{tgds}, a methodology for trajectory-guided discharge stratification in \gls{hf} that transforms structured \glspl{ehr} into compact, recency-weighted patient sequences modeled by a short-context Llama backbone. On a Swedish \gls{hf} cohort ($N=42{,}820$), \gls{tgds} outperforms  a trivial \gls{xgb} baseline, advanced baselines such as the well-known BERT architecture for \gls{ehr} sequence modeling, and Mamba backbones on the two one-year prediction tasks (clinical instability and mortality after the initial \gls{hf} diagnosis), with robust calibration. Component-by-component ablations support each design choice in \gls{tgds}: category-level tokenization, the recency-weighted short-context representation, and the Llama backbone are each independently validated. Tiny variants already surpass larger advanced baselines, highlighting the importance of an \gls{ntp} objective over model size. \Gls{tgds} also degrades gracefully under reduced clinical-concept availability and training data. The joint clinical instability and mortality predictions provide a framework for stratifying patients into four clinically meaningful risk profiles rather than standalone risk scores, which may support future discharge planning and personalized care pathways pending prospective clinical evaluation.

\section*{CRediT authorship contribution statement}
\textbf{Falk Dippel:} Writing – review \& editing, Writing – original draft, Visualization, Validation, Software, Project administration, Methodology, Investigation, Formal analysis, Data curation, Conceptualization. \textbf{Yinan Yu:} Writing – review \& editing, Visualization, Validation, Supervision, Resources, Project administration, Methodology, Funding acquisition, Formal analysis, Conceptualization. \textbf{Annika Rosengren:} Writing – review \& editing, Resources, Conceptualization. \textbf{Martin Lindgren:} Writing – review \& editing, Conceptualization. \textbf{Christina E. Lundberg:} Writing – review \& editing, Conceptualization. \textbf{Erik Aerts:} Writing – review \& editing, Conceptualization. \textbf{Martin Adiels:} Writing – review \& editing, Formal analysis, Conceptualization. \textbf{Helen Sjöland:} Writing – review \& editing, Validation, Supervision, Resources, Project administration, Methodology, Funding acquisition, Conceptualization.

\section*{Ethics statement}
The study conforms to the principles outlined in the Declaration of Helsinki. The study was approved and informed consent from the study participants was waived by the Swedish Ethical Review Authority, through the Ethics Committee of Umeå University (EPN reference: DNR 2021-02786).

\section*{Declaration of competing interest}
The authors declare no competing interest.

\section*{Acknowledgments}
This work was supported by grants from the Swedish state under an agreement concerning research and education of doctors (ALFGBG-991470 [to H.S.]), the Swedish Research Council (VRREG 2023-06421 [to H.S.]), Forte – the Swedish Research Council for Health, Working Life and Welfare (DNR 2024-00277 [to H.S.]), and Vinnova Advanced Digitalization (DNR 2024-01446 [to Y.Y. and H.S.]). The authors would like to thank the VGR AI platform for providing the essential GPU resources to conduct the experiments.

\section*{Data availability}
The data underlying this paper are not available for sharing due to participant privacy and ethical restrictions.

\section*{Code availability}
The code used in this paper is available at \url{https://github.com/dipfa/tgds-hf}.

\section*{Supplementary data}
Supplementary material related to this article can be found attached. 

\bibliographystyle{elsarticle-num}
\bibliography{references}


\appendix

\AtBeginEnvironment{tabular}{\footnotesize}
\AtBeginEnvironment{tabular*}{\footnotesize}

\captionsetup[figure]{
    name=Fig.,
    font=normal
}

\captionsetup[table]{
    font=normal
}

\glsdisablehyper 

\newacronym{gelu}{GeLU}{Gaussian error linear unit}
\newacronym{rope}{RoPE}{rotary position embedding}
\newacronym{rmsnorm}{RMSNorm}{root mean square layer normalization}

\newcommand{\cmark}{\ding{51}}
\newcommand{\xmark}{\ding{55}}

\newcommand{\tone}{\textbf{$T_1$: One-year clinical instability after initial \gls{hf} diagnosis}}
\newcommand{\ttwo}{\textbf{$T_2$: One-year mortality after initial \gls{hf} diagnosis}}

\newcommand{\chvvarone}{\textit{Variation of architecture}}
\newcommand{\chvvartwo}{\textit{Variation of tokenization}}
\newcommand{\chvvarthree}{\textit{Variation of temporal preprocessing}}
\newcommand{\chvvarfour}{\textit{Variation of model scale (context length $C$; model size $S$)}}

\glsresetall 

\section*{Supplementary material} 

\appendix

\renewcommand{\thesection}{S\arabic{section}}
\setcounter{section}{0}

\renewcommand{\thefigure}{S\arabic{figure}}
\setcounter{figure}{0}

\renewcommand{\thetable}{S\arabic{table}}
\setcounter{table}{0}

\section{Data}
This section provides additional information regarding the patient data or clinical prediction tasks. Table~\ref{stab:features} highlights the extracted clinical concepts from \glspl{ehr}. The eligibility criteria for clinical instability is shown in Table~\ref{stab:crit_instability}. Table~\ref{stab:cohort} presents the statistical properties of the \gls{hf} cohort at the initial \gls{hf} diagnosis in-hospital.

\begin{table}[!h]
    \centering
    \caption{The patient data consists of six different feature groups. \gls{atc}=Anatomical Therapeutic Chemical. \gls{icd}=International Classification of Diseases 10th revision.}
    \label{stab:features}
    \begin{tabular}{lp{5cm}}
    \toprule
    Concept & Subconcept \\ \midrule
    Demographics & Age, sex, body mass index \\
    Diagnosis (DX) & Any ICD-10 code \\
    Vital signs (VIT) & Body temperature, diastolic blood pressure, systolic blood pressure, pulse rate, respiratory rate, oxygen saturation \\
    Laboratories (LAB) & Albumin, bilirubin, blood urea nitrogen, C-reactive protein, creatinine, ferritin, fasting glucose, plasma glucose, hemoglobin, glycated hemoglobin, N-terminal pro b-type natriuretic peptide, potassium, sodium, alanine transaminase, aspartate transaminase, troponin I, troponin T, uric acid
  \\
    Procedures (PRO) & Any KVÅ code \\
    Medications (MED) & Following ATC code levels: A02B, A08, B01, B03, C, H03, R03 \\
    \bottomrule
    \end{tabular}
\end{table}

\begin{table}[!h]
    \centering
    \caption{Eligibility criteria for unplanned hospitalizations in the clinical instability prediction task.}
    \label{stab:crit_instability}
    \begin{tabular}{p{2cm}p{5.5cm}}
    \toprule
    Hospitalization type & Eligibility criteria \\ \midrule
    Unplanned \gls{hf} hospitalization & Service units: ÖS Med/Ger/Akut, Medicinklinik, Kardiologi, Medicin, Internmedicin, Specialistmedicin, MS Medicin o Akutmottagning, M210 Medicin Lidköping, Akutmedicin \& Geriatri, M360 Kardiologi, Infektion-hematologi-hud, MS Geriatrik, Geriatrik neurologi och rehabilitering, Thorax, M220 Medicin Falköping, M420 AVA, Infektion, M330 Infektion, Njurmedicin, M350 Njurmedicin, Lungmedicin, Allergologi och Palliativ medicin, M130 Lungmedicin och neurologi, M120 Hematologi, M110 Stroke och rehabilitering, Akutmedicin, M230 Medicin Mariestad, K250 Palliativ vård, M160 Medicin slutenvård, M110 Stroke, neurologi och rehabilitering, M130 Lungmedicin och medicin, Klinik för nära vård,  M365 Mag och tarm, M390 Geriatrik \\
    \makecell[l]{Unplanned non-\gls{hf} \\ hospitalization \\}
    & \makecell[l]{Surgical KVÅ codes: F, G, J, N, TF, V, XF, ZF; \\ Medical KVÅ codes: AF, AG, AP, DF, DG, DP} \\ \bottomrule
    \end{tabular}
\end{table}

\begin{table}[!h]
\centering
\caption{Descriptive statistics and outcome prevalence for the clinical prediction tasks in the \gls{hf} cohort at the initial \gls{hf} diagnosis in-hospital. IQR: interquartile range.}
\label{stab:cohort}
\begin{tabular}{ll}
\toprule
Information & Initial HF diagnosis \\  \midrule
$N$ & $42{,}820$ \\
\makecell[l]{Age (right-sided) \\ \phantom{0}} & \makecell[l]{IQR: 80 (72, 87), \\ min: 18, max: 107} \\
Sex (\% women) & 45.4\% \\
\makecell[l]{Hospitalizations \\ \phantom{0}} & \makecell[l]{IQR: 2 (1, 3), \\min: 1, max: 61} \\
\makecell[l]{All events \\ \phantom{0}} & \makecell[l]{IQR: 153 (71, 324), \\ min: 1, max: 9552} \\
\makecell[l]{DX events \\ \phantom{0}} & \makecell[l]{IQR: 9 (5, 16), \\ min: 1, max: 311} \\
\makecell[l]{VIT events \\ \phantom{0}} & \makecell[l]{IQR: 30 (0, 91), \\ min: 0, max: 2634} \\
\makecell[l]{LAB events \\ \phantom{0}} & \makecell[l]{IQR: 23 (8, 55), \\ min: 0, max: 3845} \\
\makecell[l]{MED events \\ \phantom{0}} & \makecell[l]{IQR: 63 (26, 137), \\ min: 0, max: 5746} \\
\makecell[l]{PRO events \\ \phantom{0}} & \makecell[l]{IQR: 3 (1, 7), \\ min: 0, max: 339} \\
\makecell[l]{Prevalence of \\ one-year instability} & \makecell[l]{39.7\% \\ \phantom{0}} \\
\makecell[l]{Prevalence of \\ one-year mortality} & \makecell[l]{24.8\%  \\ \phantom{0}} \\
\bottomrule
\end{tabular}
\end{table}

\section{Methods}

\subsection{Sequence model architectures} \label{ssec:sequence_models}

This section compares the different architectural design choices while Table~\ref{stab:architecture} summarizes the design choices.

\begin{table}[!h]
    \centering
    \caption{Comparison of architectural key design choices. MBERT=ModernBERT.}
    \label{stab:architecture}
    \begin{tabular}{lllll}
    \toprule 
    \makecell[l]{Model \\ \phantom{0}} & \makecell[l]{Learning \\ objective} & \makecell[l]{Position \\ encodings} & \makecell[l]{Activation \\ function} & \makecell[l]{Applied \\ normalization} \\ 
    \midrule
    BERT &  MLM & Absolute & GeLU & LayerNorm (post) \\
    XLNet & PLM & Relative & GeLU & LayerNorm (post) \\
    MBERT & MLM &  RoPE & GeGLU & LayerNorm (pre) \\
    Llama & NTP & RoPE & SwiGLU & RMSNorm (pre) \\
    Mamba & NTP & --- & SiLU & RMSNorm (pre) \\
    Mamba2 & NTP &--- & SiLU & RMSNorm (pre) \\
    \bottomrule
    \end{tabular}
\end{table}

\Gls{bert} captures bidirectional contextual information via self-attention and is pre-trained on a \gls{mlm} objective, where tokens of the input sequence are randomly masked and predicted given the surrounding context. XLNet predicts tokens autoregressively utilizing a \gls{plm} objective. In \gls{plm}, tokens are randomly permuted based on a factorization order allowing to capture more long-range dependencies beyond the fixed context provided by auto-encoders \cite{yang2019xlnet}. 
Although \gls{plm} undermines the temporal structure of \gls{ehr} sequences, this flexibility can be advantageous when capturing interactions among co-occurring clinical events. ModernBERT (hereafter referred to as MBERT) and Llama include several improvements over BERT. The absolute position embedding is replaced with \gls{rope} \cite{smsu2024roformer} capturing both the absolute position of tokens and the relative distance between tokens through rotation encoding. MBERT replaces the \gls{gelu} activation function with Gaussian error gated linear unit (GeGLU) and Llama with  a combination of GLU and Swish activation (SwiGLU) due to the apparent performance gains \cite{smshazeer2020glu}. The post-normalization of the input after passing through a Transformer sub-layer is replaced with pre-normalization to improve training stability \cite{smxiong2020layer}. Llama replaces LayerNorm with \gls{rmsnorm} \cite{smzhang2019root} reducing the computational burden by eliminating the mean calculation required in LayerNorm. For better representation learning MBERT introduces an alternating mechanism between global and local attention every three layers \cite{smwarner2024smarter}.

Mambas like Llama are pre-trained on the \gls{ntp} objective, where the next token is predicted based on only past tokens. Unlike Transformers where the attention mechanism causes quadratic training time, the Mamba architecture combines discretized \gls{ssm} with a context-aware selection mechanism that parameterizes the \gls{ssm} to learn relevant context while hardware-aware algorithms enable linear scaling with sequence length \cite{gu2023mamba}. Mamba2 improves Mamba's efficiency by introducing the state space duality which connects \gls{ssm} with attention heads and simplifies the structure of the matrix $\mathbf{A}$ in the state equation of \gls{ssm} into a scalar multiple of the identity \cite{dao2024transformers}. Both Mambas utilize a sigmoid linear unit (SiLU) activation function and \gls{rmsnorm} as pre-normalization. 

\subsection{Model theory} \label{ssec:model_theory}
This section explains the theoretical concepts behind the attention mechanism in Transformer and \gls{ssm} modeling in Mamba.

\subsubsection{Multihead attention in Transformer} \label{ssec:attention}

To calculate the attention weights,  a scaled dot-product attention is formulated as 
\begin{equation}
    A = \mathrm{Softmax}\left(\frac{Q K{^\intercal}}{\sqrt{d_k}}\right) V
\end{equation}
where $Q$, $K$, $V$ denote the query, key and value matrices and $d_k$ the dimension of keys \cite{vaswani2017attention}. The vectors $Q$, $K$, $V$ are linear projections of the input sequence  $x$ using weights as $Q=xW^Q$, $K=xW^K$, $V=xW^V$ where the weights $W^Q \in \mathbb{R}^{d_\mathrm{m} \times d_k}$, $W^K \in \mathbb{R}^{d_\mathrm{m} \times d_k}$ and $W^V \in \mathbb{R}^{d_\mathrm{m} \times d_v}$ are trainable model parameters with $d_m$ being the dimension of the embeddings and $d_v$ being the dimension of value. The $\mathrm{MHA}$ weights are calculated as

\begin{equation}
    \mathrm{MHA} = (A_i \oplus \ldots \oplus A_{n_h}) W^O
\end{equation} 
by concatenating the outputs of $n_h$ parallelized attention heads, followed by a linear projection with the trainable output matrix $W^O \in \mathbb{R}^{n_hd_v \times d_\mathrm{model}}$.

\subsubsection{State space model in Mamba}  \label{ssec:ssm}
The Mamba architecture is based on a discrete \gls{ssm} formulation paired with convolutional computation power. In general, a continuous \gls{ssm} describes the system dynamics through a state equation
\begin{equation}
    h'(t) = \mathbf{A} h(t) + \mathbf{B} x(t)
\end{equation}
using $n$-dimensional hidden states $h(t) \in \mathbb{R}^{n}$ and maps a state to an output  $y(t) \in \mathbb{R}$ through the observation equation 
\begin{equation}
    y(t) = \mathbf{C} h(t) + \mathbf{D} x(t)
\end{equation}
for a given input $x(t) \in \mathbb{R}$ at current time $t$ using system matrices $\mathbf{A}\in \mathbb{R}^{n\times n}$, $\mathbf{B}\in \mathbb{R}^{n\times 1}$, $\mathbf{C}\in \mathbb{R}^{1\times n}$, $\mathbf{D}\in \mathbb{R}$ with the common choice of $\mathbf{D}=0$. Given a discrete time step $k$, the discrete \gls{ssm} is expressed as 
\begin{equation}
    h_k = \bar{\mathbf{A}} h_{k-1} + \bar{\mathbf{B}} x_k
\end{equation}
\begin{equation}
    y_k = \bar{\mathbf{C}} h_{k}
\end{equation}
with the discretised matrix versions 
 $\bar{\mathbf{A}}=\exp(\Delta \mathbf{A})$ and $\bar{\mathbf{B}}=(\Delta \mathbf{A})^{-1} (\exp(\Delta \mathbf{A}) - \mathbf{I})$ obtained from zero-order hold discretization over a constant interval $\Delta=\interval[scaled]{t_{k-1}}{t_k}$ \cite{gu2023mamba}. Similar to multihead attention, the matrix calculation is formulated using a global convolution kernel $\bar{K}$
\begin{equation}
  y=x \ast \bar{\mathbf{K}} 
\end{equation}
with $\bar{\mathbf{K}}=(\mathbf{C}\bar{\mathbf{B}}, \ldots, \mathbf{C}\bar{\mathbf{A}}^{k}\bar{\mathbf{B}}, \ldots)$.

\subsection{Implementation} \label{ssec:implementation} 
For each clinical task, the data was divided into a development ($85\%$) and testing ($15\%$) set stratified by the outcome of interest. The development set was further split into a training ($90\%$) and validation set ($10\%$). No patient appeared in more than one set. Here, the development set was used for both pre-training the sequence model on a model-specific learning objective and fine-tuning on the relevant clinical task. Patients were consistently assigned to the same set and split across both stages. We motivate this choice by our pre-selection of only \gls{hf} patients as opposed to other, larger cohorts with more general clinical profiles \cite{li2020behrt, rasmy2021med, pang2021cehr, rupp2023exbehrt}.

Our code implementation is adapted from \cite{fallahpour2024ehrmamba}. Code was implemented using Python 3.10.12, Ubuntu 22.04, and the following major packages: torch 2.2.0, transformers 4.49.0, mamba-ssm 2.2.2, causal-conv1d 1.4.0, and xgboost 2.0.0. Data processing used 32 INTEL(R) XEON(R) PLATINUM 8568Y+ CPUs. All sequence models were trained on a single NVIDIA H200 with 50\% GPU resource allocation ($\sim$70 GB memory) and the same patient embeddings. Pre-training was set to 30 epochs with a patience of 5 epochs, fine-tuning to 10 with a patience of 2 epochs. In both stages, the optimizer choice was AdamW \cite{smloshchilov2017decoupled} with a learning rate of $5\cdot10^{-5}$ and a batch size of 32. In all dropout layers, the probability was set to $0.1$. The \gls{mlm} objective was masked with a probability of 0.15 in BERT and MBERT. For both Mambas, the size of the space state vector was set to 16, the size of the convolutional kernel to 4 and the Mamba block expansion to 2 matching the size of Transformer blocks \cite{gu2023mamba}. \Gls{xgb} was trained using the default parameters (a total of 100 trees, a maximum depth of 6 per tree, learning rate of 0.3). During fine-tuning, predicted probabilities were obtained through a two-layer feedforward classification head applied on the last hidden state of the rightmost non-padded token, and the backbone model was further trained using binary cross-entropy loss.

\subsection{Experiments}
Tables~\ref{stab:vocab}--\ref{stab:context} highlight the varying properties of the patient sequences given the respective conducted ablation. For varying model configurations (Tiny, Small, Medium), Table~\ref{stab:modelcfg} lists the hyperparameters while Table~\ref{stab:modelsize} compares the model parameter counts and \gls{gmacs}. 

\begin{table}[!h]
\centering
\caption{Ablation of vocabulary $V$: The vocabulary size is influenced by the number of bins $b$ to represent \gls{vit}, \gls{lab} and \gls{med}, as well as if \gls{dx} are represented by category ($i=3$) or diagnosis \gls{icd} codes ($i=4$).}
\label{stab:vocab}
\begin{tabular}{@{}lcccccc@{}}
\toprule
Vocabulary & Total & DX & VIT & LAB & MED & PRO \\ \midrule
\vocone & 3293 & 1343 & 36 & 108 & \phantom{0}858 & 948 \\
\voctwo & 4128 & 1343 & 66 & 198 & 1573 & 948 \\
\vocthree & 6450 & 4500 & 36 & 108 & \phantom{0}858 & 948 \\
\vocfour  & 7285 & 4500 & 66 & 198 & 1573 & 948 \\
\bottomrule
\end{tabular}
\end{table}

\begin{table}[!ht]
    \centering
    \caption{Ablation of patient's medical history $H$ with context length $C=512$: Truncation includes encounter information only within the latest admission ($t=0$) or up to $t \in \{1, 3\}$ years of prior $H$. Cutoff refers to the unprocessed $H$ determined by $C$. Aggregation averages continuous features within $w \in \{1, 2\}$ days during the encounter duration extending the longitudinal gap $\Delta$Days between the first and last event within $H$.}
    \label{stab:hist}
    \setlength{\tabcolsep}{4pt}
    \begin{tabular}{@{}lllll@{}}
    \toprule
    Dataset & Modification & \multicolumn{1}{c}{Tokens} & \multicolumn{1}{c}{Visits} & \multicolumn{1}{c}{$\Delta$Days} \\ \midrule
    \histtzero & Truncation & \phantom{0}96 (50, 179) & 1 (1, 1) & \phantom{00}6 (3, 11) \\
    \histtone & Truncation & 124 (61, 248) & 1 (1, 2) & \phantom{0}10 (4, 50) \\
    \histtthree & Truncation & 143 (68, 297) & 1 (1, 2) & \phantom{0}17 (5, 276) \\
    \histcontext & Cutoff & 153 (71, 324) & 1 (1, 3) & \phantom{0}27 (6, 542) \\
    \histwinone & Aggregation & 112 (55, 229) & 1 (1, 3) & \phantom{0}31 (6, 647) \\
    \histwintwo & Aggregation & \phantom{0}81 (42, 161) & 1 (1, 3) & \phantom{0}32 (6, 704) \\ \bottomrule
    \end{tabular}
\end{table}

\begin{table}[!h]
\centering
\caption{Ablation of context length $C$: The value of $C$ influences the number of tokens, the number of hospitalizations (visits), and the longitudinal gap $\Delta$Days between last and first event.}
\label{stab:context}
\begin{tabular}{@{}lllll@{}}
\toprule
Vocabulary & $C$ & \multicolumn{1}{c}{Tokens} & \multicolumn{1}{c}{Visits} & \multicolumn{1}{c}{$\Delta$Days} \\ \midrule
\voctwo & 128 & 128 (71, 128) & 1 (1, 2) & \phantom{0}11 (5, 46) \\
\voctwo & 256 & 153 (71, 256) & 1 (1, 2) & \phantom{0}19 (6, 280) \\
\voctwo & 512 & 153 (71, 324) & 1 (1, 3) & \phantom{0}27 (6, 542) \\
\bottomrule
\end{tabular}
\end{table}

\begin{table}[!h]
\centering
\caption{Ablation of model size $S$: All model hyperparameters are listed in the order of $\mathrm{Tiny}$/$\mathrm{Small}$/$\mathrm{Medium}$ configurations. The dimension of the hidden layers is denoted as $d_m$, the number of hidden layers as $n_m$, the number of (attention) heads $n_h$, the number of Mamba blocks as $n_b$ and the dimension of heads $d_p$. The dimension of the feed-forward layer $d_f$ was set to 512/1024/2048.}
\label{stab:modelcfg}
\setlength{\tabcolsep}{5pt}
\begin{tabular}{lccccc}
\toprule 
Model & $d_m$ & $n_m$ & $n_h$ & $n_b$ & $d_p$\\ \midrule 
BERT & 256/512/512 & 4/4/6 & 4/4/8  & --- & --- \\
XLNet & 256/512/512 & 4/4/6 & 4/4/8  & --- & --- \\
MBERT & 256/512/512 & 4/4/6 & 4/4/8 & --- & --- \\
Llama & 256/512/512 & 4/4/6 & 4/4/8  & --- & --- \\
Mamba & 256/512/512 & --- & --- & 2/6/12 & --- \\
Mamba2 & 256/512/512 & --- & 16/16/16  & 2/6/12 & 32/64/64 \\
\bottomrule
\end{tabular}
\end{table}

\begin{table}[!h]
\centering
\caption{Total model parameters in M (MParams) and \gls{gmacs} are listed in the order of $\mathrm{Tiny}$/$\mathrm{Small}$/$\mathrm{Medium}$ configurations for fine-tuned models including classifier head. \gls{gmacs} were calculated using the ptflops package (version 0.7.5).}
\label{stab:modelsize}
\setlength{\tabcolsep}{4pt}
\begin{tabular}{@{}lll@{}}
\toprule
Model & MParams & GMACs \\ \midrule
BERT & 3.4/13.6/24.1 & 13.3/57.3/96.2 \\
XLNet & 6.7/17.3/28.4 & 7.0/27.4/62.1 \\
MBERT & 10.7/17.6/32.2 & 26.2/54.1/114.2 \\
Llama & 12.9/19.7/34.4 & 25.1/53.1/113.3 \\
Mamba & 3.3/15.3/25.5 & 3.8/41.5/81.7 \\
Mamba2 & 4.3/16.9/26.6 & 3.7/40.6/79.9 \\ \bottomrule
\end{tabular}
\end{table}

\section{Additional results} \label{ssec:results}
\subsection{Additional metrics for ablation study} 
\begin{figure*}[!b]
    \centering
    \begin{subfigure}[t]{0.48\textwidth}
        \centering
        \includegraphics[width=\linewidth]{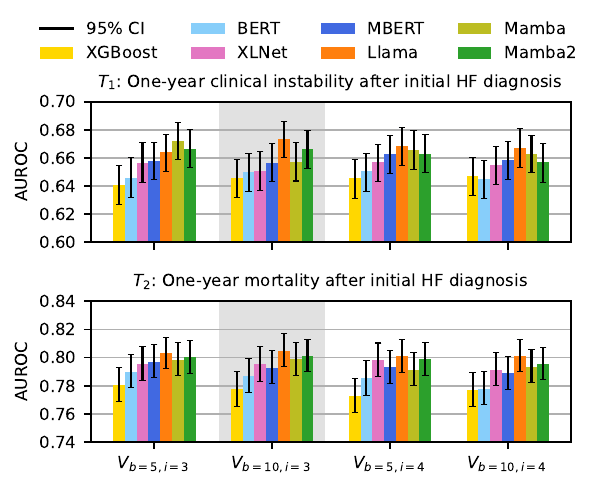}
        \caption{}
        \label{sfig:ex1_auroc}
    \end{subfigure}
    \hfill 
    \begin{subfigure}[t]{0.48\textwidth}
        \centering
        \includegraphics[width=\linewidth]{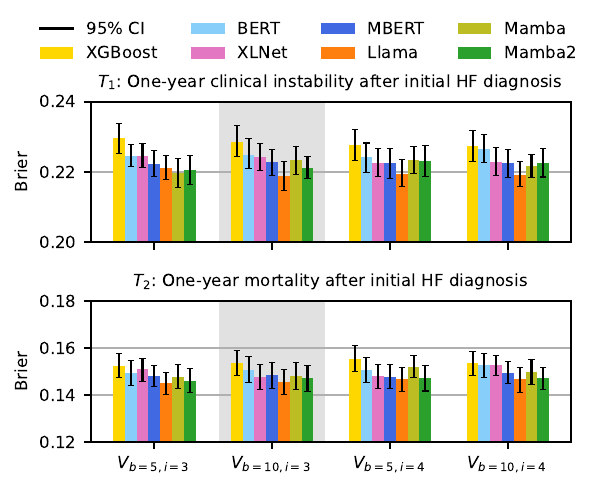}
        \caption{}
        \label{sfig:ex1_brier}
    \end{subfigure}
    \caption{Ablation of the vocabulary $V$ across the two clinical tasks $T$ evaluated by (a) \gls{auroc} ($\uparrow$)  and (b) Brier score ($\downarrow$) for $\mathrm{Medium}$-sized sequence models and $C=512$. The resolution of the vocabulary increases from lowest (left) to highest (right). Gray background highlights shared setup across all ablations.}
    \label{sfig:ex1}
\end{figure*}

\begin{figure*}[!h]
    \centering
    \begin{subfigure}[t]{0.48\textwidth}
        \centering
        \includegraphics[width=\linewidth]{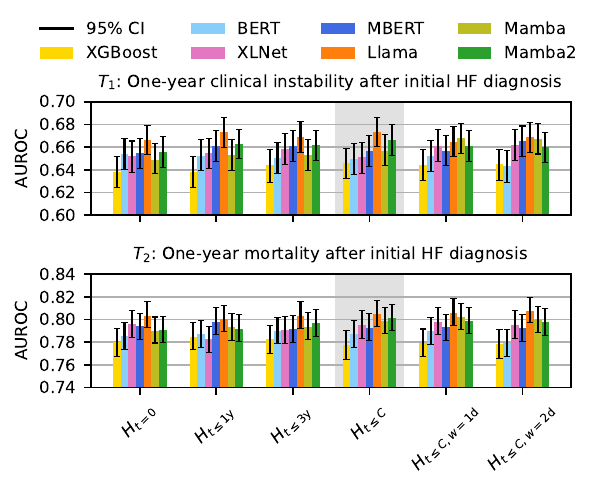}
        \caption{}
        \label{sfig:ex2_auroc}
    \end{subfigure}
    \hfill 
    \begin{subfigure}[t]{0.48\textwidth}
        \centering
        \includegraphics[width=\linewidth]{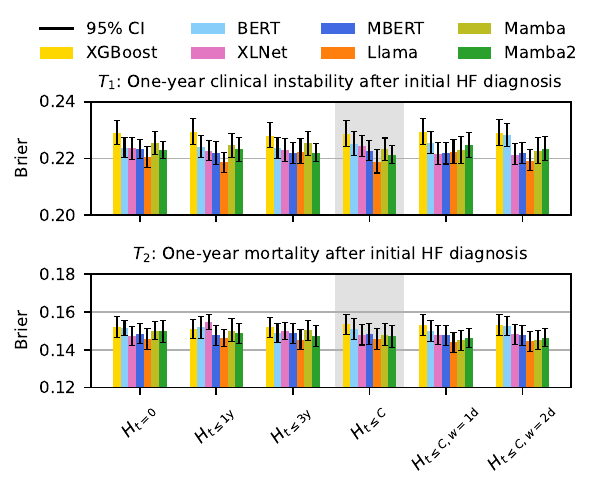}
        \caption{}
        \label{sfig:ex2_brier}
    \end{subfigure}
    \caption{Ablation of the patient history through temporal modifications: The historical record is extended from the latest available admission (left) to a prolonged context (right). All modifications are evaluated by (a) \gls{auroc} ($\uparrow$) and (b) Brier score ($\downarrow$) using $\mathrm{Medium}$-sized sequence models and $C=512$. Gray background highlights shared setup across all ablations.}
    \label{sfig:ex2}
\end{figure*}

\begin{figure*}[!h]
    \centering
    \begin{subfigure}[t]{0.48\textwidth}
        \centering
        \includegraphics[width=\linewidth]{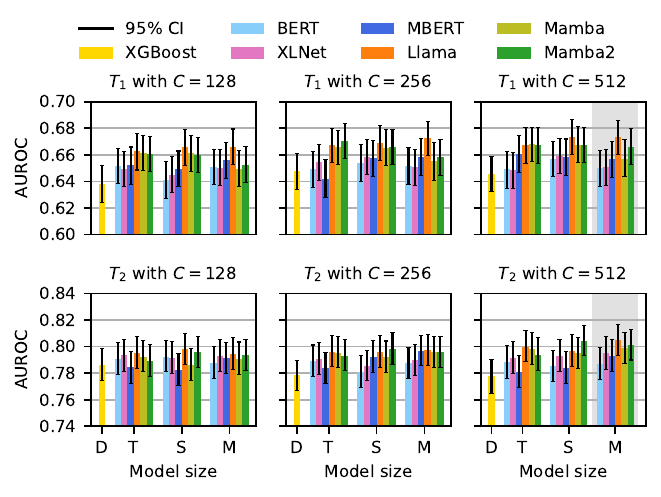}
        \caption{}
        \label{sfig:ex3_auroc}
    \end{subfigure}
    \hfill 
    \begin{subfigure}[t]{0.48\textwidth}
        \centering
        \includegraphics[width=\linewidth]{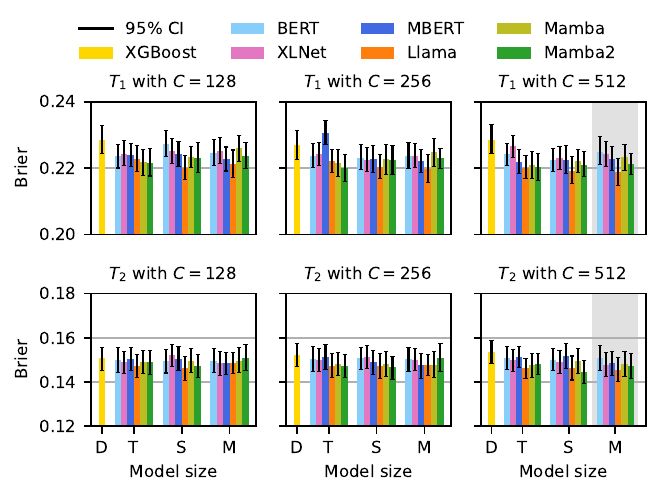}
        \caption{}
        \label{sfig:ex3_brier}
    \end{subfigure}
    \caption{Ablation of the context length $C$ and model size evaluated by (a) \gls{auroc} ($\uparrow$) and (b) Brier score ($\downarrow$) across the two clinical tasks. Within each $C$ the size of the sequence models is sorted by $\mathrm{\underline{T}iny}, \mathrm{\underline{S}mall}, \mathrm{\underline{M}edium}$ and compared with the $\mathrm{\underline{D}efault}$ \gls{xgb} configuration. Gray background highlights shared setup across all ablations.}
    \label{sfig:ex3}
\end{figure*}

\begin{figure*}[!h]
    \centering
    \begin{subfigure}[t]{0.48\textwidth}
        \centering
        \includegraphics[width=\linewidth]{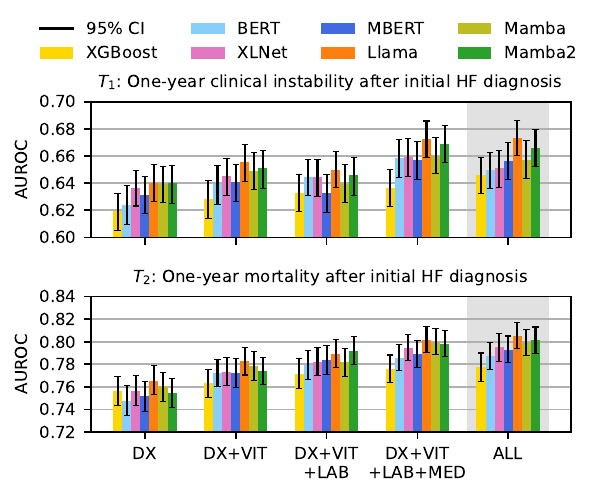}
        \caption{}
        \label{sfig:ex4_auroc}
    \end{subfigure}
    \hfill 
    \begin{subfigure}[t]{0.48\textwidth}
        \centering
        \includegraphics[width=\linewidth]{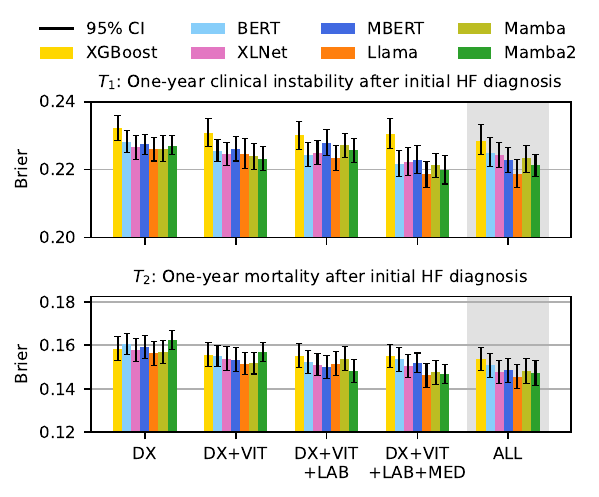}
        \caption{}
        \label{sfig:ex4_brier}
    \end{subfigure}
    \caption{Incremental augmentation of \gls{dx} with  clinical concepts until all, including \gls{pro}, are added. Each augmentation is evaluated by (a) \gls{auroc} ($\uparrow$) and (b) Brier score ($\downarrow$) using $\mathrm{Medium}$-sized sequence models and $C=512$. Gray background highlights shared setup across all ablations.}
    \label{sfig:ex4}
\end{figure*}

\begin{figure*}[!h]
    \centering
    \begin{subfigure}[t]{0.48\textwidth}
        \centering
        \includegraphics[width=\linewidth]{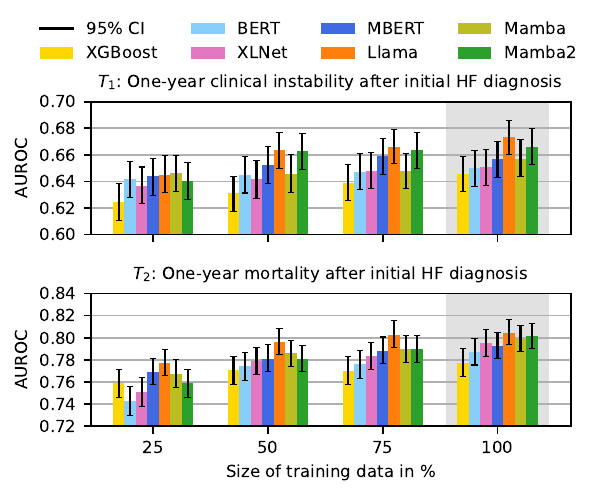}
        \caption{}
        \label{sfig:ex5_auroc}
    \end{subfigure}
    \hfill 
    \begin{subfigure}[t]{0.48\textwidth}
        \centering
        \includegraphics[width=\linewidth]{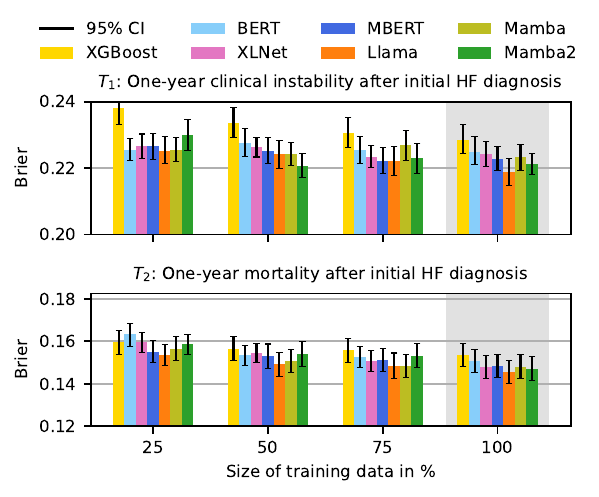}
        \caption{}
        \label{sfig:ex5_brier}
    \end{subfigure}
    \caption{Ablation of different training sizes for $\mathrm{Medium}$-sized sequence models and $C=512$ evaluated by (a) \gls{auroc} ($\uparrow$) and (b) Brier score ($\downarrow$). Gray background highlights shared setup across all ablations.}
    \label{sfig:ex5}
\end{figure*}

This section presents additional results, primarily the evaluation of \gls{auroc} and Brier score. Table~\ref{stab:key_findings_add} highlights the main findings of this paper. For all conducted ablations the \gls{auroc} and Brier score are visualized in Figs.~\ref{sfig:ex1}--\ref{sfig:ex5} and numerical results of all metrics, including \gls{auprc}, are presented in Tables~\ref{stab:ex1_auprc}--~\ref{stab:ex5_brier}. 

\subsection{Profiling} 
\begin{figure*}[!ht]
    \centering
    \begin{subfigure}[t]{0.48\textwidth}
        \centering
        \includegraphics[width=\linewidth]{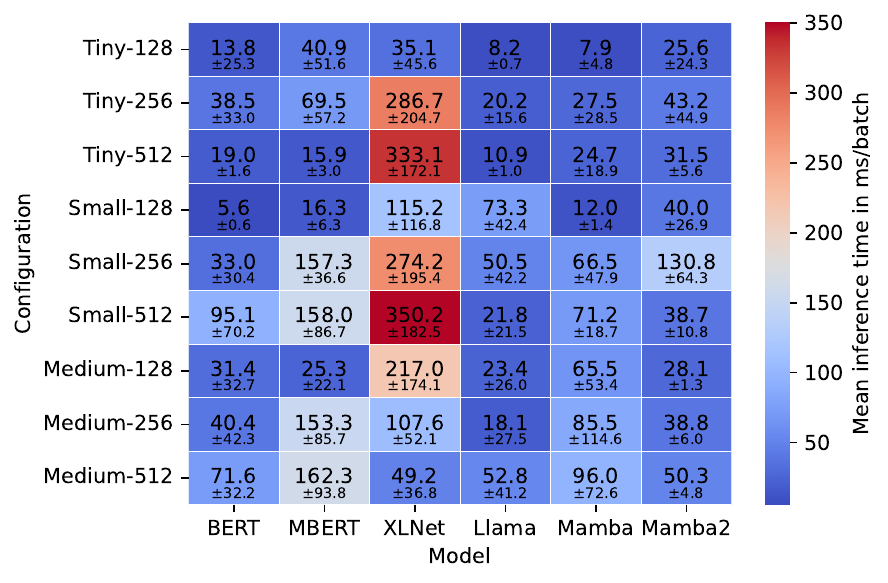}
        \caption{Mean inference runtime in pre-training setting.}
        \label{sfig:inf_time_pt}
    \end{subfigure}
    \hfill 
    \begin{subfigure}[t]{0.48\textwidth}
        \centering
        \includegraphics[width=\linewidth]{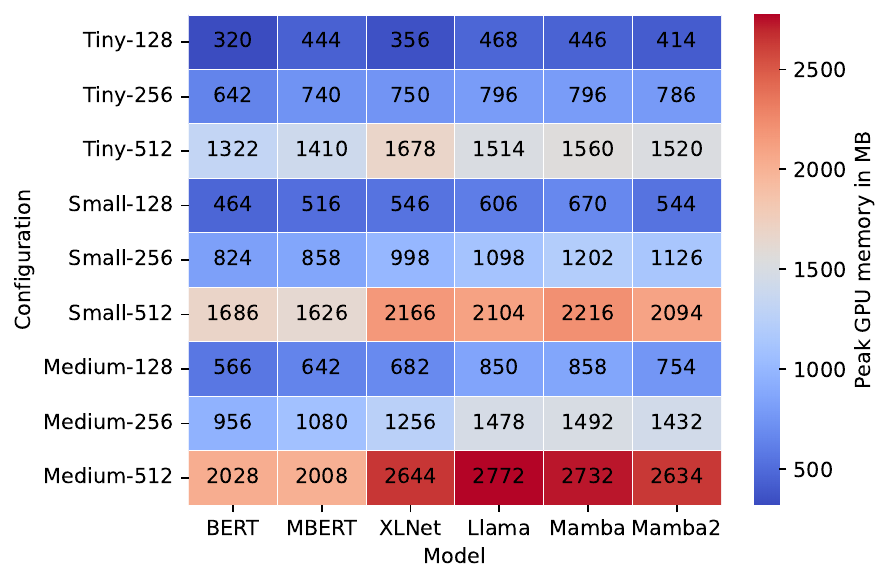}
        \caption{Mean inference runtime in fine-tuning setting.}
        \label{sfig:inf_time_ft}
    \end{subfigure}
    \begin{subfigure}[t]{0.48\textwidth}
        \centering
        \includegraphics[width=\linewidth]{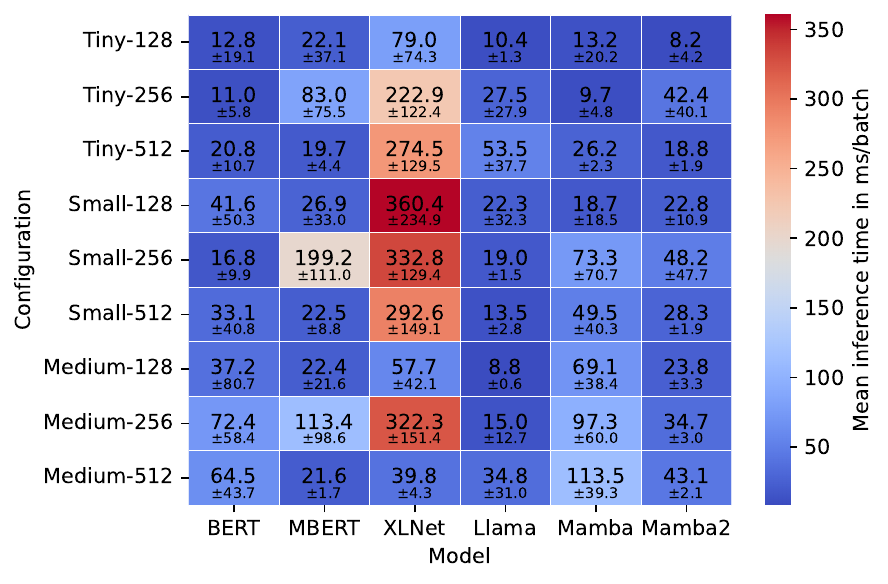}
        \caption{Inference peak GPU memory usage in pre-training setting.}
        \label{sfig:inf_gpu_pt}
    \end{subfigure}
    \hfill 
    \begin{subfigure}[t]{0.48\textwidth}
        \centering
        \includegraphics[width=\linewidth]{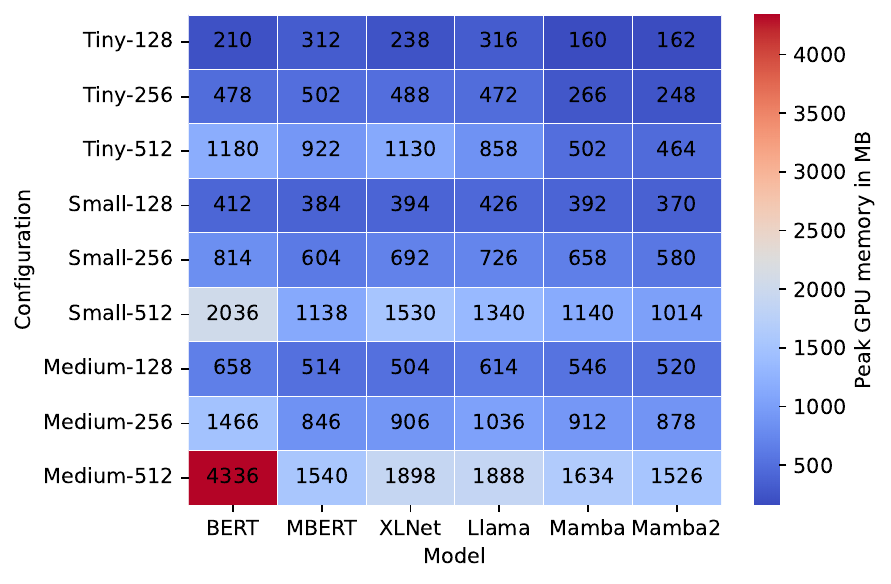}
        \caption{Inference peak GPU memory usage in fine-tuning setting.}
        \label{sfig:inf_gpu_ft}
    \end{subfigure}
    \caption{Inference measurements for runtime and peak GPU memory usage across 10 runs with a batch size of 32, stratified by models during pre-training and fine-tuning setting, with measurements based on the prediction of one-year mortality after the initial \gls{hf} diagnosis.}
    \label{sfig:inference}
\end{figure*}

Regarding profiling, Table~\ref{stab:train_time} provides an overview of the training time for both pre-training and fine-tune stages, whereas Fig.~\ref{sfig:inference} highlights the runtime and peak GPU memory usage in inference mode.

\subsection{Internal hospital validation} 
\begin{figure*}
    \centering
    \includegraphics[width=\linewidth, scale=0.75]{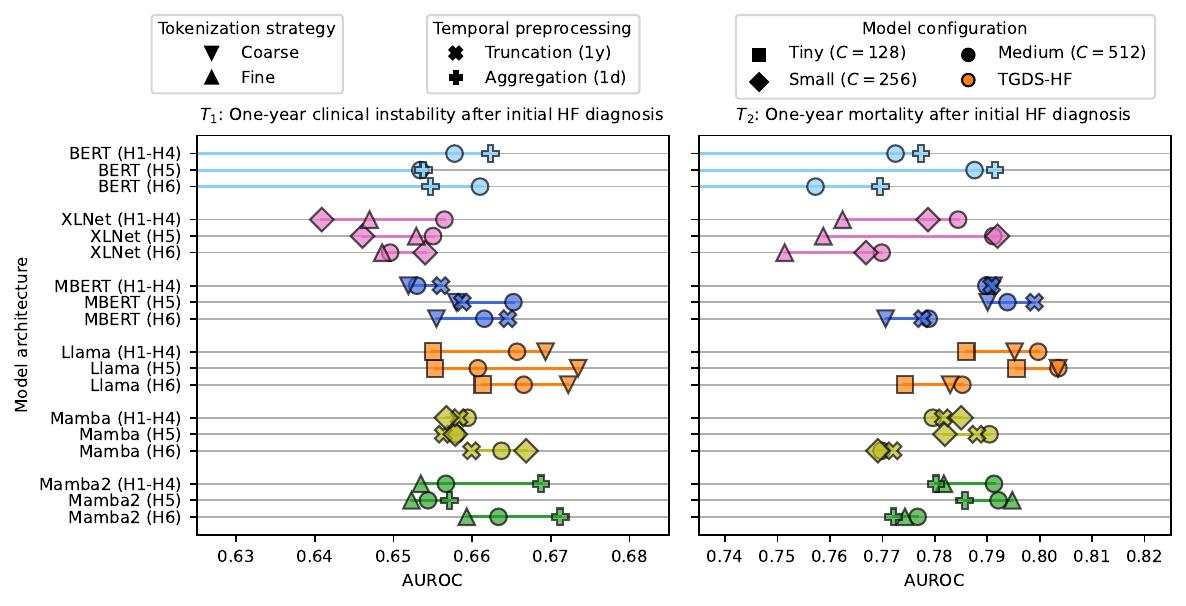}
    \caption{Evaluation of \gls{auroc} ($\uparrow$) in cross-hospital validation grouped by model architecture and stratified by independent hospital sets.}
    \label{sfig:eval_chv_auroc}
\end{figure*}

\begin{figure*}
    \centering
    \includegraphics[width=\linewidth, scale=0.75]{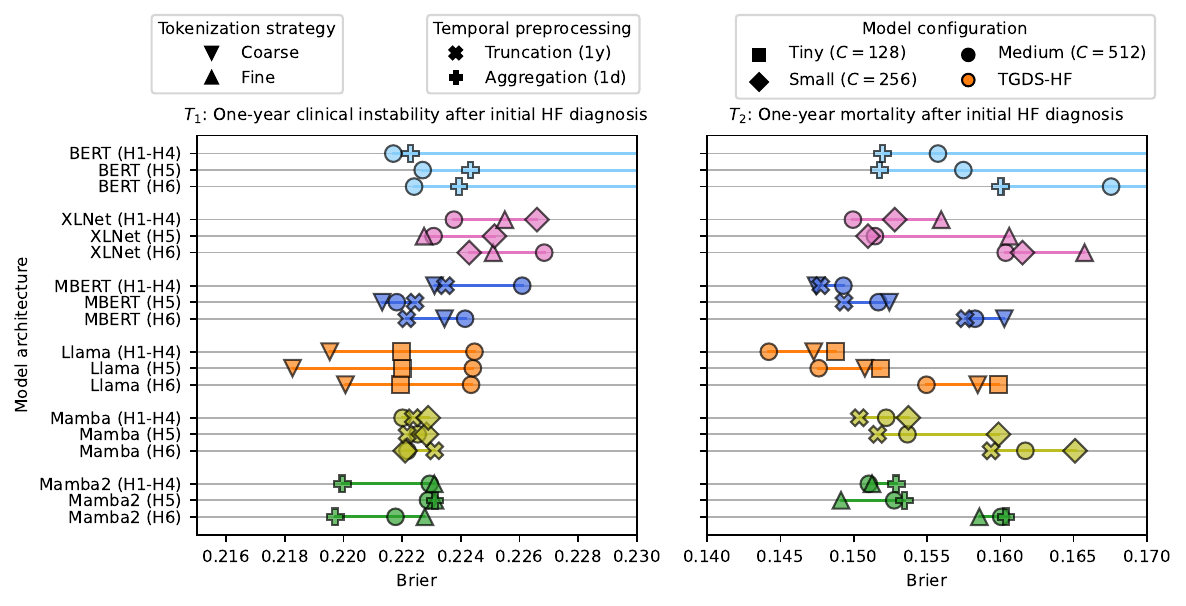}
    \caption{Evaluation of Brier score ($\downarrow$) in cross-hospital validation grouped by model architecture and stratified by independent hospital sets.}
    \label{sfig:eval_chv_brier}
\end{figure*}

The descriptive statistics for the \gls{shv} and \gls{chv} are provided in Table~\ref{stab:val_hosp_stats}. Table~\ref{stab:eval_shv} shows the numerical metrics (\gls{auprc}, \gls{auroc}, Brier score) for \gls{shv}, whereas Tables~\ref{stab:eval_chv_auprc}--\ref{stab:eval_chv_brier} show them for \gls{chv} with corresponding design choices, and also visualized in Fig.~\ref{sfig:eval_chv_auroc}--\ref{sfig:eval_chv_brier}.

\begin{table*}[h!]
    \centering
    \caption{Main result of \gls{tgds} vs.\ established baselines for $C=512$, mirroring Table~2 in the manuscript for \gls{auroc} and Brier score.}
    \label{stab:key_findings_add}
    
    \begin{subtable}{0.8\textwidth}
    \centering
    \caption{\textbf{Best} \gls{auroc} (95\% \gls{ci}) performances per task are highlighted in bold.}
    \label{stab:key_findings_auroc}
    \setlength{\tabcolsep}{4pt}
    \begin{tabular}{@{}llllrll@{}}
    \toprule
    Model & Objective & Temporal & Model size & \multicolumn{1}{l}{GMACs} & \multicolumn{1}{c}{$T_1$} & \multicolumn{1}{c}{$T_2$} \\ \midrule
    \multicolumn{7}{@{}l}{\textit{Established baselines}} \\
     XGBoost & ---& Cutoff & --- & --- & 0.646 \ci{(0.632--0.659)} & 0.778 \ci{(0.765--0.790)} \\
     BERT & MLM & Cutoff & Medium & 96.2 & 0.650 \ci{(0.636--0.663)} & 0.787 \ci{(0.775--0.799)} \\
     XLNet & PLM & Cutoff & Medium & 62.1 & 0.651 \ci{(0.637--0.664)} & 0.795 \ci{(0.783--0.808)} \\
    MBERT & MLM & Cutoff & Medium & 114.2 & 0.656 \ci{(0.643--0.670)} & 0.793 \ci{(0.781--0.805)} \\
    \multicolumn{7}{@{}l}{\textit{\gls{tgds} (proposed)}} \\
    \textbf{TGDS-HF (Llama)} & NTP & Cutoff & Medium & 113.3 & \textbf{0.673} \ci{(0.660--0.686)} & 0.805 \ci{(0.794--0.817)} \\
    \multicolumn{7}{@{}l}{\textit{Alternative NTP backbones (ablation A3)}} \\
    Mamba & NTP & Cutoff & Medium & 81.7 & 0.657 \ci{(0.644--0.671)} & 0.799 \ci{(0.787--0.811)} \\
     Mamba2 & NTP & Cutoff & Medium & 79.9 & 0.666 \ci{(0.653-0.680)} & 0.801 \ci{(0.790--0.813)} \\
    \multicolumn{7}{@{}l}{\textit{Task-specific refinement: daily aggregation for the mortality task} ($T_2$)} \\
     \textbf{TGDS-HF (Llama, agg.)} & NTP & Agg. (1d) & Medium & 113.3 & 0.665 \ci{(0.651--0.678)} & \textbf{0.806} \ci{(0.795--0.818)} \\
     Mamba (agg.) & NTP & Agg. (1d) & Medium & 81.7 & 0.668 \ci{(0.654--0.681)} & 0.802 \ci{(0.791--0.814)} \\
    \multicolumn{7}{@{}l}{\textit{Tiny variants of \gls{tgds} already surpass Medium baselines}} \\
     \textbf{TGDS-HF (Llama, Tiny)} & NTP & Cutoff & Tiny & 25.1 & 0.668 \ci{(0.654--0.681)} & 0.800 \ci{(0.789--0.812)} \\
     Mamba (Tiny) & NTP & Cutoff & Tiny & 3.8 & 0.668 \ci{(0.655--0.681)} & 0.798 \ci{(0.786--0.810)} \\ \bottomrule
    \end{tabular}
    \end{subtable}

    \vspace{0.8cm}

    \begin{subtable}{0.8\textwidth}
    \centering
    \caption{\textbf{Best} Brier score (95\% \gls{ci}) performances per task are highlighted in bold.}
    \label{stab:key_findings_brier}
    \setlength{\tabcolsep}{4pt}
    \begin{tabular}{@{}llllrll@{}}
    \toprule
    Model & Objective & Temporal & Model size & \multicolumn{1}{l}{GMACs} & \multicolumn{1}{c}{$T_1$} & \multicolumn{1}{c}{$T_2$} \\ \midrule
    \multicolumn{7}{@{}l}{\textit{Established baselines}} \\
     XGBoost & ---& Cutoff & --- & --- & 0.229 \ci{(0.224--0.233)} & 0.153 \ci{(0.148--0.159)} \\
     BERT & MLM & Cutoff & Medium & 96.2 & 0.225 \ci{(0.221--0.229)} & 0.151 \ci{(0.145--0.156)} \\
     XLNet & PLM & Cutoff & Medium & 62.1 & 0.224 \ci{(0.221--0.228)} & 0.148 \ci{(0.142--0.153)} \\
    MBERT & MLM & Cutoff & Medium & 114.2 & 0.223 \ci{(0.219--0.226)} & 0.148 \ci{(0.143--0.154)} \\
    \multicolumn{7}{@{}l}{\textit{\gls{tgds} (proposed)}} \\
    \textbf{TGDS-HF (Llama)} & NTP & Cutoff & Medium & 113.3 & \textbf{0.219} \ci{(0.215--0.223)} & 0.146 \ci{(0.140--0.151)} \\
    \multicolumn{7}{@{}l}{\textit{Alternative NTP backbones (ablation A3)}} \\
    Mamba & NTP & Cutoff & Medium & 81.7 & 0.223 \ci{(0.219--0.227)} & 0.148 \ci{(0.142--0.154)} \\
     Mamba2 & NTP & Cutoff & Medium & 79.9 & 0.221 \ci{(0.218--0.224)} & 0.147 \ci{(0.141--0.153)} \\
    \multicolumn{7}{@{}l}{\textit{Task-specific refinement: daily aggregation for the mortality task} ($T_2$)} \\
     \textbf{TGDS-HF (Llama, agg.)} & NTP & Agg. (1d) & Medium & 113.3 & 0.222 \ci{(0.218--0.227)} & \textbf{0.144} \ci{(0.138--0.149)} \\
     Mamba (agg.) & NTP & Agg. (1d) & Medium & 81.7 & 0.223 \ci{(0.218--0.228)} & 0.145 \ci{(0.140--0.150)} \\
    \multicolumn{7}{@{}l}{\textit{Tiny variants of \gls{tgds} already surpass Medium baselines}} \\
     \textbf{TGDS-HF (Llama, Tiny)} & NTP & Cutoff & Tiny & 25.1 & 0.220 \ci{(0.217--0.224)} & 0.146 \ci{(0.141--0.151)} \\
     Mamba (Tiny) & NTP & Cutoff & Tiny & 3.8 & 0.221 \ci{(0.217--0.225)} & 0.148 \ci{(0.142--0.153)} \\ \bottomrule
    \end{tabular}
    \end{subtable}
\end{table*}


\begin{table*}[!h]
\centering
\caption{Ablation of the vocabulary $V$ across the two clinical tasks $T$ evaluated by \gls{auprc} ($\uparrow$) for $\mathrm{Medium}$-sized sequence models and $C=512$. Gray background highlights shared setup across all ablations.}
\label{stab:ex1_auprc}
\setlength{\tabcolsep}{2pt}
\begin{tabular}{@{}l|l|l|l|l|l|l|l@{}}
\toprule
\multicolumn{1}{c}{Vocabulary} & \multicolumn{1}{c}{XGBoost} & \multicolumn{1}{c}{BERT} & \multicolumn{1}{c}{XLNet} & \multicolumn{1}{c}{ModernBERT} & \multicolumn{1}{c}{Llama} & \multicolumn{1}{c}{Mamba} & \multicolumn{1}{c}{Mamba2} \\ \midrule
\multicolumn{7}{c}{\tone{}} \\
\vocone{} & 0.515 (0.494--0.535) & 0.525 (0.506--0.547) & 0.539 (0.518--0.558) & 0.536 (0.516--0.558) & 0.547 (0.527--0.567) & 0.552 (0.532--0.573) & 0.543 (0.522--0.564) \\
\rowcolor{gray!20}
\voctwo{} & 0.519 (0.497--0.540) & 0.535 (0.516--0.556) & 0.529 (0.508--0.549) & 0.534 (0.513--0.556) & 0.555 (0.535--0.575) & 0.543 (0.523--0.565) & 0.542 (0.521--0.562) \\
\vocthree{} & 0.521 (0.500--0.541) & 0.535 (0.515--0.555) & 0.544 (0.524--0.564) & 0.546 (0.526--0.567) & 0.549 (0.529--0.570) & 0.545 (0.526--0.566) & 0.545 (0.525--0.567) \\
\vocfour{} & 0.523 (0.502--0.544) & 0.527 (0.507--0.547) & 0.535 (0.515--0.555) & 0.533 (0.514--0.553) & 0.550 (0.529--0.570) & 0.544 (0.523--0.564) & 0.536 (0.516--0.556) \\
\multicolumn{7}{c}{\ttwo{}} \\
\vocone{} & 0.525 (0.501--0.552) & 0.549 (0.525--0.574) & 0.549 (0.523--0.575) & 0.561 (0.537--0.586) & 0.563 (0.537--0.589) & 0.563 (0.537--0.588) & 0.566 (0.542--0.591) \\
\rowcolor{gray!20}
\voctwo{} & 0.525 (0.500--0.550) & 0.540 (0.516--0.564) & 0.558 (0.533--0.584) & 0.557 (0.531--0.583) & 0.574 (0.550--0.599) & 0.559 (0.534--0.585) & 0.572 (0.547--0.596) \\
\vocthree{} & 0.513 (0.489--0.538) & 0.539 (0.514--0.565) & 0.552 (0.526--0.579) & 0.560 (0.535--0.584) & 0.562 (0.536--0.586) & 0.545 (0.521--0.570) & 0.562 (0.536--0.586) \\
\vocfour{} & 0.524 (0.498--0.549) & 0.522 (0.496--0.545) & 0.540 (0.514--0.566) & 0.546 (0.521--0.570) & 0.565 (0.541--0.592) & 0.553 (0.528--0.579) & 0.559 (0.533--0.585) \\
\bottomrule
\end{tabular}
\end{table*}

\begin{table*}[!h]
\centering
\caption{Ablation of the vocabulary $V$ across the two clinical tasks $T$ evaluated by \gls{auroc} ($\uparrow$) for $\mathrm{Medium}$-sized sequence models and $C=512$. Gray background highlights shared setup across all ablations.}
\label{stab:ex1_auroc}
\setlength{\tabcolsep}{2pt}
\begin{tabular}{@{}l|l|l|l|l|l|l|l@{}}
\toprule
\multicolumn{1}{c}{Vocabulary} & \multicolumn{1}{c}{XGBoost} & \multicolumn{1}{c}{BERT} & \multicolumn{1}{c}{XLNet} & \multicolumn{1}{c}{ModernBERT} & \multicolumn{1}{c}{Llama} & \multicolumn{1}{c}{Mamba} & \multicolumn{1}{c}{Mamba2} \\
\midrule
\multicolumn{7}{c}{\tone{}} \\
\vocone{} & 0.641 (0.627--0.655) & 0.646 (0.632--0.660) & 0.657 (0.643--0.671) & 0.658 (0.645--0.671) & 0.664 (0.650--0.677) & 0.672 (0.659--0.685) & 0.667 (0.653--0.680) \\
\rowcolor{gray!20}
\voctwo{} & 0.646 (0.632--0.659) & 0.650 (0.636--0.663) & 0.651 (0.637--0.664) & 0.656 (0.643--0.670) & 0.673 (0.660--0.686) & 0.657 (0.644--0.671) & 0.666 (0.653--0.680) \\
\vocthree{} & 0.645 (0.631--0.659) & 0.650 (0.636--0.663) & 0.657 (0.643--0.670) & 0.662 (0.649--0.676) & 0.669 (0.655--0.681) & 0.666 (0.652--0.679) & 0.663 (0.650--0.677) \\
\vocfour{} & 0.647 (0.633--0.661) & 0.645 (0.631--0.658) & 0.655 (0.641--0.668) & 0.658 (0.645--0.671) & 0.667 (0.654--0.681) & 0.663 (0.650--0.676) & 0.657 (0.642--0.671) \\
\multicolumn{7}{c}{\ttwo{}} \\
\vocone{} & 0.781 (0.769--0.793) & 0.790 (0.779--0.802) & 0.795 (0.784--0.808) & 0.797 (0.786--0.809) & 0.803 (0.792--0.815) & 0.798 (0.787--0.810) & 0.801 (0.789--0.812) \\
\rowcolor{gray!20}
\voctwo{} & 0.778 (0.765--0.790) & 0.787 (0.775--0.799) & 0.795 (0.783--0.808) & 0.793 (0.781--0.805) & 0.805 (0.794--0.817) & 0.799 (0.787--0.811) & 0.801 (0.790--0.813) \\
\vocthree{} & 0.772 (0.761--0.785) & 0.785 (0.773--0.798) & 0.798 (0.786--0.810) & 0.793 (0.782--0.805) & 0.801 (0.789--0.813) & 0.791 (0.780--0.804) & 0.799 (0.788--0.811) \\
\vocfour{} & 0.777 (0.765--0.789) & 0.778 (0.767--0.790) & 0.791 (0.780--0.803) & 0.789 (0.777--0.801) & 0.801 (0.790--0.813) & 0.793 (0.782--0.806) & 0.795 (0.784--0.807) \\
\bottomrule
\end{tabular}
\end{table*}

\begin{table*}[!h]
\centering
\caption{Ablation of the vocabulary $V$ across the two clinical tasks $T$ evaluated by Brier score ($\downarrow$) for $\mathrm{Medium}$-sized sequence models and $C=512$. Gray background highlights shared setup across all ablations.}
\label{stab:ex1_brier}
\setlength{\tabcolsep}{2pt}
\begin{tabular}{@{}l|l|l|l|l|l|l|l@{}}
\toprule
\multicolumn{1}{c}{Vocabulary} & \multicolumn{1}{c}{XGBoost} & \multicolumn{1}{c}{BERT} & \multicolumn{1}{c}{XLNet} & \multicolumn{1}{c}{ModernBERT} & \multicolumn{1}{c}{Llama} & \multicolumn{1}{c}{Mamba} & \multicolumn{1}{c}{Mamba2} \\
\multicolumn{7}{c}{\tone{}} \\
\vocone{} & 0.230 (0.225--0.234) & 0.225 (0.221--0.228) & 0.225 (0.221--0.228) & 0.222 (0.219--0.226) & 0.221 (0.218--0.225) & 0.220 (0.216--0.224) & 0.221 (0.216--0.225) \\
\rowcolor{gray!20}
\voctwo{} & 0.229 (0.224--0.233) & 0.225 (0.221--0.229) & 0.224 (0.221--0.228) & 0.223 (0.219--0.226) & 0.219 (0.215--0.223) & 0.223 (0.219--0.227) & 0.221 (0.218--0.224) \\
\vocthree{} & 0.228 (0.223--0.232) & 0.224 (0.220--0.228) & 0.223 (0.219--0.227) & 0.223 (0.218--0.227) & 0.220 (0.216--0.224) & 0.223 (0.220--0.227) & 0.223 (0.219--0.228) \\
\vocfour{} & 0.227 (0.223--0.232) & 0.227 (0.223--0.231) & 0.223 (0.219--0.227) & 0.223 (0.219--0.226) & 0.219 (0.216--0.223) & 0.222 (0.218--0.225) & 0.223 (0.219--0.227) \\
\multicolumn{7}{c}{\ttwo{}} \\
\vocone{} & 0.153 (0.147--0.158) & 0.149 (0.144--0.155) & 0.151 (0.146--0.156) & 0.148 (0.144--0.153) & 0.145 (0.140--0.150) & 0.148 (0.143--0.153) & 0.146 (0.141--0.151) \\
\rowcolor{gray!20}
\voctwo{} & 0.153 (0.148--0.159) & 0.151 (0.145--0.156) & 0.148 (0.142--0.153) & 0.148 (0.143--0.154) & 0.146 (0.140--0.151) & 0.148 (0.142--0.154) & 0.147 (0.141--0.153) \\
\vocthree{} & 0.155 (0.150--0.161) & 0.151 (0.145--0.156) & 0.148 (0.143--0.153) & 0.148 (0.143--0.153) & 0.147 (0.141--0.152) & 0.152 (0.147--0.157) & 0.147 (0.142--0.153) \\
\vocfour{} & 0.153 (0.148--0.159) & 0.153 (0.148--0.158) & 0.153 (0.148--0.157) & 0.149 (0.145--0.154) & 0.147 (0.141--0.152) & 0.150 (0.144--0.155) & 0.147 (0.142--0.152) \\
\bottomrule
\end{tabular}
\end{table*}

\begin{table*}[!h]
\centering
\caption{Ablation of the history $H$ across the two clinical tasks $T$ evaluated by \gls{auprc} ($\uparrow$) for $\mathrm{Medium}$-sized sequence models and $C=512$. Gray background highlights shared setup across all ablations.}
\label{stab:ex2_auprc}
\setlength{\tabcolsep}{2pt}
\begin{tabular}{@{}l|l|l|l|l|l|l|l@{}}
\toprule
\multicolumn{1}{c}{History} & \multicolumn{1}{c}{XGBoost} & \multicolumn{1}{c}{BERT} & \multicolumn{1}{c}{XLNet} & \multicolumn{1}{c}{ModernBERT} & \multicolumn{1}{c}{Llama} & \multicolumn{1}{c}{Mamba} & \multicolumn{1}{c}{Mamba2} \\ \midrule
\multicolumn{7}{c}{\tone{}} \\
\histtzero{} & 0.515 (0.495--0.536) & 0.531 (0.510--0.552) & 0.531 (0.511--0.552) & 0.536 (0.516--0.557) & 0.541 (0.521--0.562) & 0.526 (0.506--0.546) & 0.528 (0.508--0.549) \\
\histtone{} & 0.518 (0.498--0.539) & 0.535 (0.515--0.555) & 0.535 (0.514--0.555) & 0.542 (0.521--0.561) & 0.557 (0.537--0.576) & 0.533 (0.513--0.553) & 0.536 (0.516--0.557) \\
\histtthree{} & 0.523 (0.502--0.545) & 0.536 (0.515--0.556) & 0.541 (0.519--0.561) & 0.543 (0.522--0.563) & 0.555 (0.535--0.576) & 0.534 (0.513--0.554) & 0.541 (0.520--0.560) \\
\rowcolor{gray!20}
\histcontext{} & 0.519 (0.497--0.540) & 0.535 (0.516--0.556) & 0.529 (0.508--0.549) & 0.534 (0.513--0.556) & 0.555 (0.535--0.575) & 0.543 (0.523--0.565) & 0.542 (0.521--0.562) \\
\histwinone{} & 0.514 (0.494--0.536) & 0.536 (0.515--0.557) & 0.543 (0.522--0.563) & 0.534 (0.514--0.554) & 0.549 (0.529--0.570) & 0.543 (0.523--0.564) & 0.533 (0.513--0.553) \\
\histwintwo{} & 0.515 (0.495--0.537) & 0.518 (0.496--0.538) & 0.543 (0.523--0.563) & 0.543 (0.521--0.563) & 0.551 (0.530--0.572) & 0.546 (0.526--0.566) & 0.536 (0.516--0.556) \\
\bottomrule
\end{tabular}
\end{table*}

\begin{table*}[!h]
\centering
\caption{Ablation of the history $H$ across the two clinical tasks $T$ evaluated by \gls{auroc} ($\uparrow$) for $\mathrm{Medium}$-sized sequence models and $C=512$. Gray background highlights shared setup across all ablations.}
\label{stab:ex2_auroc}
\setlength{\tabcolsep}{2pt}
\begin{tabular}{@{}l|l|l|l|l|l|l|l@{}}
\toprule
\multicolumn{1}{c}{History} & \multicolumn{1}{c}{XGBoost} & \multicolumn{1}{c}{BERT} & \multicolumn{1}{c}{XLNet} & \multicolumn{1}{c}{ModernBERT} & \multicolumn{1}{c}{Llama} & \multicolumn{1}{c}{Mamba} & \multicolumn{1}{c}{Mamba2} \\ \midrule
\multicolumn{7}{c}{\tone{}} \\
\histtzero{} & 0.638 (0.625--0.652) & 0.654 (0.640--0.668) & 0.652 (0.638--0.665) & 0.655 (0.641--0.667) & 0.666 (0.653--0.679) & 0.649 (0.636--0.663) & 0.656 (0.642--0.669) \\
\histtone{} & 0.638 (0.624--0.652) & 0.652 (0.639--0.666) & 0.655 (0.641--0.668) & 0.661 (0.647--0.675) & 0.673 (0.659--0.686) & 0.653 (0.639--0.666) & 0.662 (0.650--0.676) \\
\histtthree{} & 0.644 (0.629--0.658) & 0.650 (0.637--0.664) & 0.658 (0.644--0.672) & 0.661 (0.647--0.674) & 0.669 (0.655--0.682) & 0.653 (0.640--0.667) & 0.661 (0.648--0.675) \\
\rowcolor{gray!20}
\histcontext{} & 0.646 (0.632--0.659) & 0.650 (0.636--0.663) & 0.651 (0.637--0.664) & 0.656 (0.643--0.670) & 0.673 (0.660--0.686) & 0.657 (0.644--0.671) & 0.666 (0.653--0.680) \\
\histwinone{} & 0.644 (0.630--0.658) & 0.652 (0.638--0.666) & 0.661 (0.647--0.675) & 0.657 (0.643--0.671) & 0.665 (0.651--0.678) & 0.668 (0.654--0.681) & 0.661 (0.647--0.674) \\
\histwintwo{} & 0.645 (0.630--0.658) & 0.643 (0.629--0.657) & 0.662 (0.649--0.675) & 0.665 (0.651--0.679) & 0.669 (0.655--0.681) & 0.667 (0.654--0.681) & 0.660 (0.647--0.673) \\
\bottomrule
\end{tabular}
\end{table*}

\begin{table*}[!h]
\centering
\caption{Ablation of the history $H$ across the two clinical tasks $T$ evaluated by Brier score ($\downarrow$) for $\mathrm{Medium}$-sized sequence models and $C=512$. Gray background highlights shared setup across all ablations.}
\label{stab:ex2_brier}
\setlength{\tabcolsep}{2pt}
\begin{tabular}{@{}l|l|l|l|l|l|l|l@{}}
\toprule
\multicolumn{1}{c}{History} & \multicolumn{1}{c}{XGBoost} & \multicolumn{1}{c}{BERT} & \multicolumn{1}{c}{XLNet} & \multicolumn{1}{c}{ModernBERT} & \multicolumn{1}{c}{Llama} & \multicolumn{1}{c}{Mamba} & \multicolumn{1}{c}{Mamba2} \\ \midrule
\multicolumn{7}{c}{\tone{}} \\
\histtzero{} & 0.229 (0.225--0.233) & 0.224 (0.220--0.227) & 0.224 (0.220--0.227) & 0.223 (0.220--0.227) & 0.220 (0.217--0.224) & 0.226 (0.222--0.230) & 0.223 (0.220--0.226) \\
\histtone{} & 0.229 (0.225--0.234) & 0.224 (0.220--0.228) & 0.223 (0.219--0.226) & 0.222 (0.218--0.226) & 0.219 (0.215--0.222) & 0.225 (0.220--0.229) & 0.223 (0.219--0.227) \\
\histtthree{} & 0.228 (0.224--0.233) & 0.224 (0.220--0.227) & 0.223 (0.219--0.227) & 0.222 (0.218--0.226) & 0.222 (0.218--0.227) & 0.225 (0.221--0.229) & 0.222 (0.219--0.225) \\
\rowcolor{gray!20}
\histcontext{} & 0.229 (0.224--0.233) & 0.225 (0.221--0.229) & 0.224 (0.221--0.228) & 0.223 (0.219--0.226) & 0.219 (0.215--0.223) & 0.223 (0.219--0.227) & 0.221 (0.218--0.224) \\
\histwinone{} & 0.229 (0.225--0.234) & 0.226 (0.222--0.230) & 0.222 (0.218--0.226) & 0.222 (0.218--0.226) & 0.222 (0.218--0.227) & 0.223 (0.218--0.228) & 0.225 (0.220--0.229) \\
\histwintwo{} & 0.229 (0.225--0.234) & 0.228 (0.224--0.232) & 0.221 (0.218--0.225) & 0.222 (0.218--0.226) & 0.219 (0.216--0.223) & 0.223 (0.218--0.227) & 0.223 (0.219--0.228) \\
\bottomrule
\end{tabular}
\end{table*}

\begin{table*}[!h]
\centering
\caption{Ablation of the context length $C$ and model size $S$ across the two clinical tasks $T$ evaluated by \gls{auprc} ($\uparrow$). Here, T, S, M denote the $\mathrm{Tiny}$, $\mathrm{Small}$, $\mathrm{Medium}$ model sizes. XGBoost's default configuration is merged with size T for readability. Gray background highlights shared setup across all ablations.}
\label{stab:ex3_auprc}
\setlength{\tabcolsep}{2pt}
\begin{tabular}{@{}l|l|l|l|l|l|l|l|l@{}}
\toprule
\multicolumn{1}{c}{$C$} & \multicolumn{1}{c}{$S$} & \multicolumn{1}{c}{XGBoost} & \multicolumn{1}{c}{BERT} & \multicolumn{1}{c}{XLNet} & \multicolumn{1}{c}{ModernBERT} & \multicolumn{1}{c}{Llama} & \multicolumn{1}{c}{Mamba} & \multicolumn{1}{c}{Mamba2} \\ \midrule
\multicolumn{8}{c}{\tone{}} \\
128 & T & 0.517 (0.496--0.538) & 0.526 (0.505--0.547) & 0.529 (0.508--0.550) & 0.530 (0.509--0.551) & 0.542 (0.522--0.563) & 0.538 (0.517--0.559) & 0.539 (0.518--0.560) \\
128 & S & - & 0.511 (0.491--0.533) & 0.525 (0.504--0.548) & 0.528 (0.507--0.549) & 0.543 (0.523--0.563) & 0.539 (0.518--0.560) & 0.538 (0.517--0.561) \\
128 & M & - & 0.533 (0.511--0.555) & 0.529 (0.507--0.551) & 0.536 (0.515--0.558) & 0.543 (0.523--0.563) & 0.529 (0.508--0.550) & 0.531 (0.510--0.551) \\
256 & T & 0.522 (0.501--0.542) & 0.531 (0.512--0.551) & 0.534 (0.513--0.556) & 0.520 (0.500--0.542) & 0.551 (0.530--0.572) & 0.542 (0.522--0.562) & 0.545 (0.526--0.567) \\
256 & S & - & 0.536 (0.515--0.556) & 0.532 (0.512--0.553) & 0.539 (0.519--0.561) & 0.551 (0.531--0.572) & 0.543 (0.523--0.564) & 0.543 (0.522--0.564) \\
256 & M & - & 0.527 (0.507--0.547) & 0.530 (0.508--0.551) & 0.540 (0.520--0.562) & 0.554 (0.534--0.575) & 0.537 (0.516--0.559) & 0.537 (0.516--0.558) \\
\bottomrule
\end{tabular}
\end{table*}

\begin{table*}[!h]
\centering
\caption{Ablation of the context length $C$ and model size $S$ across the two clinical tasks $T$ evaluated by \gls{auroc} ($\uparrow$). Here, T, S, M denote the $\mathrm{Tiny}$, $\mathrm{Small}$, $\mathrm{Medium}$ model sizes. XGBoost's default configuration is merged with size T for readability. Gray background highlights shared setup across all ablations.}
\label{stab:ex3_auroc}
\setlength{\tabcolsep}{2pt}
\begin{tabular}{@{}l|l|l|l|l|l|l|l|l@{}}
\toprule
\multicolumn{1}{c}{$C$} & \multicolumn{1}{c}{$S$} & \multicolumn{1}{c}{XGBoost} & \multicolumn{1}{c}{BERT} & \multicolumn{1}{c}{XLNet} & \multicolumn{1}{c}{ModernBERT} & \multicolumn{1}{c}{Llama} & \multicolumn{1}{c}{Mamba} & \multicolumn{1}{c}{Mamba2} \\ \midrule
\multicolumn{8}{c}{\tone{}} \\
128 & T & 0.638 (0.624--0.652) & 0.652 (0.638--0.665) & 0.649 (0.636--0.662) & 0.652 (0.638--0.666) & 0.663 (0.649--0.676) & 0.662 (0.648--0.675) & 0.661 (0.647--0.674) \\
128 & S & - & 0.641 (0.627--0.655) & 0.645 (0.631--0.659) & 0.649 (0.636--0.663) & 0.666 (0.652--0.679) & 0.661 (0.648--0.674) & 0.660 (0.646--0.673) \\
128 & M & - & 0.651 (0.637--0.664) & 0.650 (0.637--0.664) & 0.656 (0.642--0.670) & 0.666 (0.652--0.679) & 0.650 (0.636--0.664) & 0.652 (0.639--0.666) \\
256 & T & 0.648 (0.634--0.661) & 0.650 (0.635--0.663) & 0.655 (0.641--0.668) & 0.642 (0.628--0.656) & 0.668 (0.654--0.680) & 0.666 (0.652--0.678) & 0.670 (0.657--0.683) \\
256 & S & - & 0.654 (0.640--0.667) & 0.658 (0.645--0.672) & 0.657 (0.644--0.671) & 0.669 (0.656--0.682) & 0.665 (0.652--0.679) & 0.666 (0.652--0.679) \\
256 & M & - & 0.651 (0.638--0.665) & 0.651 (0.637--0.664) & 0.658 (0.645--0.673) & 0.673 (0.659--0.685) & 0.655 (0.641--0.669) & 0.658 (0.644--0.672) \\
\bottomrule
\end{tabular}
\end{table*}

\begin{table*}[!h]
\centering
\caption{Ablation of the context length $C$ and model size $S$ across the two clinical tasks $T$ evaluated by Brier score ($\downarrow$). Here, T, S, M denote the $\mathrm{Tiny}$, $\mathrm{Small}$, $\mathrm{Medium}$ model sizes. XGBoost's default configuration is merged with size T for readability. Gray background highlights shared setup across all ablations.}
\setlength{\tabcolsep}{2pt}
\label{stab:ex3_brier}
\begin{tabular}{@{}l|l|l|l|l|l|l|l|l@{}}
\toprule
\multicolumn{1}{c}{$C$} & \multicolumn{1}{c}{$S$} & \multicolumn{1}{c}{XGBoost} & \multicolumn{1}{c}{BERT} & \multicolumn{1}{c}{XLNet} & \multicolumn{1}{c}{ModernBERT} & \multicolumn{1}{c}{Llama} & \multicolumn{1}{c}{Mamba} & \multicolumn{1}{c}{Mamba2} \\ \midrule
\multicolumn{8}{c}{\tone{}} \\
128 & T & 0.228 (0.224--0.233) & 0.223 (0.220--0.227) & 0.224 (0.221--0.228) & 0.224 (0.220--0.227) & 0.223 (0.219--0.227) & 0.222 (0.218--0.226) & 0.222 (0.218--0.226) \\
128 & S & - & 0.227 (0.223--0.231) & 0.225 (0.221--0.229) & 0.224 (0.220--0.228) & 0.220 (0.216--0.224) & 0.223 (0.220--0.227) & 0.223 (0.219--0.228) \\
128 & M & - & 0.224 (0.221--0.229) & 0.225 (0.221--0.229) & 0.223 (0.219--0.226) & 0.221 (0.217--0.225) & 0.226 (0.222--0.230) & 0.224 (0.220--0.228) \\
256 & T & 0.227 (0.223--0.231) & 0.224 (0.220--0.227) & 0.224 (0.221--0.228) & 0.231 (0.227--0.234) & 0.222 (0.219--0.226) & 0.221 (0.217--0.226) & 0.220 (0.216--0.224) \\
256 & S & - & 0.223 (0.219--0.227) & 0.222 (0.219--0.226) & 0.223 (0.219--0.227) & 0.220 (0.217--0.224) & 0.223 (0.218--0.227) & 0.222 (0.218--0.227) \\
256 & M & - & 0.224 (0.220--0.228) & 0.224 (0.220--0.227) & 0.222 (0.219--0.226) & 0.220 (0.216--0.224) & 0.225 (0.221--0.229) & 0.223 (0.220--0.226) \\
\bottomrule
\end{tabular}
\end{table*}

\begin{table*}[!h]
\centering
\caption{Ablation of incremental concept augmentation across the two clinical tasks $T$ evaluated by \gls{auprc} ($\uparrow$) for $\mathrm{Medium}$-sized sequence models and $C=512$. Here, feature group $F_1$ refers to \gls{dx}, $F_2$ to \gls{dx}+\gls{vit}, $F_3$ to \gls{dx}+\gls{vit}+\gls{lab}, $F_4$ to \gls{dx}+\gls{vit}+\gls{lab}+\gls{med} and $F_5$ to $F_4$ inclusive \gls{pro}. Gray background highlights shared setup across all ablations.}
\label{stab:ex4_auprc}
\setlength{\tabcolsep}{2pt}
\begin{tabular}{@{}l|l|l|l|l|l|l|l@{}}
\toprule
\multicolumn{1}{c}{Feature} & \multicolumn{1}{c}{XGBoost} & \multicolumn{1}{c}{BERT} & \multicolumn{1}{c}{XLNet} & \multicolumn{1}{c}{ModernBERT} & \multicolumn{1}{c}{Llama} & \multicolumn{1}{c}{Mamba} & \multicolumn{1}{c}{Mamba2} \\ \midrule
\multicolumn{7}{c}{\tone{}} \\
$F_1$ & 0.487 (0.467--0.507) & 0.502 (0.483--0.523) & 0.518 (0.498--0.538) & 0.510 (0.490--0.530) & 0.518 (0.498--0.538) & 0.514 (0.494--0.534) & 0.520 (0.500--0.540) \\
$F_2$ & 0.498 (0.478--0.519) & 0.513 (0.492--0.534) & 0.518 (0.497--0.538) & 0.517 (0.497--0.538) & 0.537 (0.516--0.557) & 0.526 (0.505--0.546) & 0.531 (0.509--0.551) \\
$F_3$ & 0.506 (0.486--0.526) & 0.521 (0.500--0.541) & 0.523 (0.502--0.545) & 0.512 (0.491--0.532) & 0.527 (0.507--0.549) & 0.519 (0.498--0.540) & 0.521 (0.502--0.542) \\
$F_4$ & 0.515 (0.494--0.534) & 0.537 (0.517--0.557) & 0.543 (0.523--0.564) & 0.537 (0.516--0.558) & 0.553 (0.533--0.574) & 0.538 (0.516--0.558) & 0.552 (0.533--0.572) \\
\rowcolor{gray!20}
ALL & 0.519 (0.497--0.540) & 0.535 (0.516--0.556) & 0.529 (0.508--0.549) & 0.534 (0.513--0.556) & 0.555 (0.535--0.575) & 0.543 (0.523--0.565) & 0.542 (0.521--0.562) \\
\multicolumn{7}{c}{\ttwo{}} \\
\bottomrule
\end{tabular}
\end{table*}

\begin{table*}[!h]
\centering
\caption{Ablation of incremental concept augmentation across the two clinical tasks $T$ evaluated by \gls{auroc} ($\uparrow$) for $\mathrm{Medium}$-sized sequence models and $C=512$. Here, $F_1$ refers to \gls{dx}, $F_2$ to \gls{dx}+\gls{vit}, $F_3$ to \gls{dx}+\gls{vit}+\gls{lab}, $F_4$ to \gls{dx}+\gls{vit}+\gls{lab}+\gls{med} and $F_5$ to $F_4$ inclusive \gls{pro}. Gray background highlights shared setup across all ablations.}
\label{stab:ex4_auroc}
\setlength{\tabcolsep}{2pt}
\begin{tabular}{@{}l|l|l|l|l|l|l|l@{}}
\toprule
\multicolumn{1}{c}{Feature} & \multicolumn{1}{c}{XGBoost} & \multicolumn{1}{c}{BERT} & \multicolumn{1}{c}{XLNet} & \multicolumn{1}{c}{ModernBERT} & \multicolumn{1}{c}{Llama} & \multicolumn{1}{c}{Mamba} & \multicolumn{1}{c}{Mamba2} \\ \midrule
\multicolumn{7}{c}{\tone{}} \\
$F_1$ & 0.619 (0.605--0.633) & 0.624 (0.610--0.638) & 0.637 (0.623--0.649) & 0.631 (0.617--0.645) & 0.640 (0.626--0.654) & 0.639 (0.626--0.652) & 0.640 (0.625--0.653) \\
$F_2$ & 0.628 (0.614--0.642) & 0.639 (0.624--0.653) & 0.645 (0.631--0.658) & 0.641 (0.626--0.654) & 0.655 (0.642--0.668) & 0.649 (0.635--0.662) & 0.651 (0.636--0.664) \\
$F_3$ & 0.633 (0.619--0.646) & 0.644 (0.631--0.658) & 0.644 (0.630--0.658) & 0.633 (0.618--0.646) & 0.650 (0.637--0.664) & 0.639 (0.625--0.654) & 0.646 (0.631--0.659) \\
$F_4$ & 0.636 (0.623--0.650) & 0.658 (0.644--0.672) & 0.659 (0.645--0.673) & 0.657 (0.642--0.671) & 0.672 (0.659--0.686) & 0.660 (0.647--0.674) & 0.669 (0.655--0.682) \\
\rowcolor{gray!20}
ALL & 0.646 (0.632--0.659) & 0.650 (0.636--0.663) & 0.651 (0.637--0.664) & 0.656 (0.643--0.670) & 0.673 (0.660--0.686) & 0.657 (0.644--0.671) & 0.666 (0.653--0.680) \\
\multicolumn{7}{c}{\ttwo{}} \\
\bottomrule
\end{tabular}
\end{table*}

\begin{table*}[!h]
\centering
\caption{Ablation of incremental concept augmentation across the two clinical tasks $T$ evaluated by Brier score ($\downarrow$) for $\mathrm{Medium}$-sized sequence models and $C=512$. Here, $F_1$ refers to \gls{dx}, $F_2$ to \gls{dx}+\gls{vit}, $F_3$ to \gls{dx}+\gls{vit}+\gls{lab}, $F_4$ to \gls{dx}+\gls{vit}+\gls{lab}+\gls{med} and $F_5$ to $F_4$ inclusive \gls{pro}. Gray background highlights shared setup across all ablations.}
\label{stab:ex4_brier}
\setlength{\tabcolsep}{2pt}
\begin{tabular}{@{}l|l|l|l|l|l|l|l@{}}
\toprule
Feature & \multicolumn{1}{c}{XGBoost} & \multicolumn{1}{c}{BERT} & \multicolumn{1}{c}{XLNet} & \multicolumn{1}{c}{ModernBERT} & \multicolumn{1}{c}{Llama} & \multicolumn{1}{c}{Mamba} & \multicolumn{1}{c}{Mamba2} \\ \midrule
\multicolumn{7}{c}{\tone{}} \\
$F_1$ & 0.232 (0.228--0.236) & 0.228 (0.225--0.232) & 0.227 (0.223--0.230) & 0.228 (0.225--0.230) & 0.226 (0.223--0.230) & 0.226 (0.222--0.230) & 0.227 (0.224--0.230) \\
$F_2$ & 0.231 (0.227--0.235) & 0.225 (0.222--0.229) & 0.225 (0.221--0.228) & 0.226 (0.223--0.230) & 0.225 (0.220--0.229) & 0.224 (0.220--0.228) & 0.223 (0.220--0.227) \\
$F_3$ & 0.230 (0.226--0.234) & 0.224 (0.221--0.228) & 0.225 (0.222--0.228) & 0.228 (0.224--0.232) & 0.223 (0.220--0.227) & 0.227 (0.224--0.231) & 0.226 (0.222--0.229) \\
$F_4$ & 0.230 (0.226--0.235) & 0.222 (0.218--0.226) & 0.222 (0.218--0.227) & 0.223 (0.219--0.227) & 0.219 (0.215--0.222) & 0.221 (0.218--0.225) & 0.220 (0.216--0.224) \\
\rowcolor{gray!20}
ALL & 0.229 (0.224--0.233) & 0.225 (0.221--0.229) & 0.224 (0.221--0.228) & 0.223 (0.219--0.226) & 0.219 (0.215--0.223) & 0.223 (0.219--0.227) & 0.221 (0.218--0.224) \\
\multicolumn{7}{c}{\ttwo{}} \\
\bottomrule
\end{tabular}
\end{table*}

\begin{table*}[!h]
\centering
\caption{Ablation of varying training set ratios across the two clinical tasks $T$ evaluated by \gls{auprc} ($\uparrow$) for $\mathrm{Medium}$-sized sequence models and $C=512$. Gray background highlights shared setup across all ablations.}
\label{stab:ex5_auprc}
\setlength{\tabcolsep}{2pt}
\begin{tabular}{@{}l|l|l|l|l|l|l|l@{}}
\toprule
\multicolumn{1}{c}{Ratio} & \multicolumn{1}{c}{XGBoost} & \multicolumn{1}{c}{BERT} & \multicolumn{1}{c}{XLNet} & \multicolumn{1}{c}{ModernBERT} & \multicolumn{1}{c}{Llama} & \multicolumn{1}{c}{Mamba} & \multicolumn{1}{c}{Mamba2} \\ \midrule
\multicolumn{7}{c}{\tone{}} \\
$25\%$ & 0.502 (0.482--0.522) & 0.516 (0.496--0.537) & 0.523 (0.503--0.544) & 0.522 (0.501--0.541) & 0.521 (0.501--0.542) & 0.528 (0.508--0.550) & 0.514 (0.493--0.534) \\
$50\%$ & 0.510 (0.492--0.531) & 0.531 (0.511--0.551) & 0.522 (0.502--0.542) & 0.535 (0.515--0.556) & 0.545 (0.525--0.566) & 0.526 (0.506--0.546) & 0.543 (0.521--0.562) \\
$75\%$ & 0.515 (0.495--0.536) & 0.527 (0.507--0.546) & 0.529 (0.508--0.549) & 0.539 (0.518--0.560) & 0.547 (0.526--0.568) & 0.523 (0.503--0.545) & 0.541 (0.521--0.561) \\
\rowcolor{gray!20}
$100\%$ & 0.519 (0.497--0.540) & 0.535 (0.516--0.556) & 0.529 (0.508--0.549) & 0.534 (0.513--0.556) & 0.555 (0.535--0.575) & 0.543 (0.523--0.565) & 0.542 (0.521--0.562) \\
\multicolumn{7}{c}{\ttwo{}} \\
$25\%$ & 0.506 (0.481--0.533) & 0.473 (0.448--0.499) & 0.475 (0.450--0.500) & 0.521 (0.497--0.547) & 0.518 (0.493--0.544) & 0.510 (0.485--0.535) & 0.483 (0.457--0.507) \\
$50\%$ & 0.508 (0.481--0.534) & 0.523 (0.498--0.548) & 0.521 (0.496--0.546) & 0.534 (0.508--0.561) & 0.564 (0.539--0.589) & 0.539 (0.511--0.563) & 0.528 (0.503--0.553) \\
$75\%$ & 0.516 (0.492--0.541) & 0.525 (0.499--0.550) & 0.539 (0.513--0.565) & 0.545 (0.520--0.571) & 0.574 (0.548--0.600) & 0.552 (0.525--0.577) & 0.553 (0.529--0.577) \\
\rowcolor{gray!20}
$100\%$ & 0.525 (0.500--0.550) & 0.540 (0.516--0.564) & 0.558 (0.533--0.584) & 0.557 (0.531--0.583) & 0.574 (0.550--0.599) & 0.559 (0.534--0.585) & 0.572 (0.547--0.596) \\
\bottomrule
\end{tabular}
\end{table*}

\begin{table*}[!h]
\centering
\caption{Ablation of varying training set ratios across the two clinical tasks $T$ evaluated by \gls{auroc} ($\uparrow$) for $\mathrm{Medium}$-sized sequence models and $C=512$. Gray background highlights shared setup across all ablations.}
\label{stab:ex5_auroc}
\setlength{\tabcolsep}{2pt}
\begin{tabular}{@{}l|l|l|l|l|l|l|l@{}}
\toprule
\multicolumn{1}{c}{Ratio} & \multicolumn{1}{c}{XGBoost} & \multicolumn{1}{c}{BERT} & \multicolumn{1}{c}{XLNet} & \multicolumn{1}{c}{ModernBERT} & \multicolumn{1}{c}{Llama} & \multicolumn{1}{c}{Mamba} & \multicolumn{1}{c}{Mamba2} \\ \midrule
\multicolumn{7}{c}{\tone{}} \\
$25\%$ & 0.624 (0.610--0.638) & 0.641 (0.628--0.655) & 0.637 (0.624--0.651) & 0.644 (0.629--0.657) & 0.645 (0.631--0.660) & 0.646 (0.633--0.660) & 0.640 (0.626--0.654) \\
$50\%$ & 0.631 (0.617--0.644) & 0.645 (0.631--0.659) & 0.641 (0.627--0.656) & 0.652 (0.639--0.666) & 0.663 (0.650--0.677) & 0.646 (0.632--0.660) & 0.663 (0.649--0.676) \\
$75\%$ & 0.639 (0.625--0.653) & 0.647 (0.634--0.661) & 0.648 (0.634--0.662) & 0.659 (0.645--0.672) & 0.666 (0.653--0.679) & 0.647 (0.635--0.661) & 0.663 (0.650--0.677) \\
\rowcolor{gray!20}
$100\%$ & 0.646 (0.632--0.659) & 0.650 (0.636--0.663) & 0.651 (0.637--0.664) & 0.656 (0.643--0.670) & 0.673 (0.660--0.686) & 0.657 (0.644--0.671) & 0.666 (0.653--0.680) \\
\multicolumn{7}{c}{\ttwo{}} \\
$25\%$ & 0.759 (0.746--0.772) & 0.743 (0.729--0.756) & 0.751 (0.738--0.764) & 0.769 (0.757--0.782) & 0.777 (0.765--0.790) & 0.767 (0.755--0.780) & 0.759 (0.746--0.771) \\
$50\%$ & 0.771 (0.757--0.783) & 0.774 (0.762--0.786) & 0.779 (0.767--0.791) & 0.781 (0.770--0.794) & 0.796 (0.784--0.808) & 0.786 (0.774--0.798) & 0.781 (0.769--0.793) \\
$75\%$ & 0.770 (0.757--0.783) & 0.776 (0.763--0.789) & 0.783 (0.771--0.796) & 0.788 (0.777--0.801) & 0.803 (0.791--0.815) & 0.790 (0.778--0.802) & 0.790 (0.778--0.802) \\
\rowcolor{gray!20}
$100\%$ & 0.778 (0.765--0.790) & 0.787 (0.775--0.799) & 0.795 (0.783--0.808) & 0.793 (0.781--0.805) & 0.805 (0.794--0.817) & 0.799 (0.787--0.811) & 0.801 (0.790--0.813) \\
\bottomrule
\end{tabular}
\end{table*}

\begin{table*}[!h]
\centering
\caption{Ablation of varying training set ratios across the two clinical tasks $T$ evaluated by Brier score ($\downarrow$) for $\mathrm{Medium}$-sized sequence models and $C=512$. Gray background highlights shared setup across all ablations.}
\label{stab:ex5_brier}
\setlength{\tabcolsep}{2pt}
\begin{tabular}{@{}l|l|l|l|l|l|l|l@{}}
\toprule
\multicolumn{1}{c}{Ratio} & \multicolumn{1}{c}{XGBoost} & \multicolumn{1}{c}{BERT} & \multicolumn{1}{c}{XLNet} & \multicolumn{1}{c}{ModernBERT} & \multicolumn{1}{c}{Llama} & \multicolumn{1}{c}{Mamba} & \multicolumn{1}{c}{Mamba2} \\ \midrule
\multicolumn{7}{c}{\tone{}} \\
$25\%$ & 0.238 (0.233--0.243) & 0.225 (0.222--0.229) & 0.227 (0.223--0.230) & 0.227 (0.223--0.231) & 0.225 (0.221--0.230) & 0.226 (0.222--0.229) & 0.230 (0.225--0.235) \\
$50\%$ & 0.234 (0.229--0.238) & 0.228 (0.224--0.232) & 0.226 (0.223--0.229) & 0.225 (0.221--0.229) & 0.224 (0.220--0.228) & 0.224 (0.221--0.228) & 0.221 (0.217--0.224) \\
$75\%$ & 0.231 (0.226--0.235) & 0.225 (0.221--0.229) & 0.223 (0.220--0.227) & 0.222 (0.218--0.226) & 0.222 (0.218--0.226) & 0.227 (0.222--0.231) & 0.223 (0.218--0.227) \\
\rowcolor{gray!20}
$100\%$ & 0.229 (0.224--0.233) & 0.225 (0.221--0.229) & 0.224 (0.221--0.228) & 0.223 (0.219--0.226) & 0.219 (0.215--0.223) & 0.223 (0.219--0.227) & 0.221 (0.218--0.224) \\
\multicolumn{7}{c}{\ttwo{}} \\
$25\%$ & 0.160 (0.154--0.165) & 0.163 (0.158--0.169) & 0.159 (0.155--0.164) & 0.155 (0.150--0.160) & 0.153 (0.149--0.158) & 0.157 (0.151--0.162) & 0.159 (0.154--0.163) \\
$50\%$ & 0.156 (0.151--0.162) & 0.153 (0.149--0.158) & 0.154 (0.150--0.159) & 0.153 (0.147--0.159) & 0.149 (0.143--0.155) & 0.151 (0.145--0.156) & 0.154 (0.148--0.160) \\
$75\%$ & 0.156 (0.150--0.161) & 0.153 (0.148--0.157) & 0.151 (0.146--0.156) & 0.151 (0.146--0.156) & 0.148 (0.143--0.154) & 0.148 (0.143--0.154) & 0.153 (0.147--0.159) \\
\rowcolor{gray!20}
$100\%$ & 0.153 (0.148--0.159) & 0.151 (0.145--0.156) & 0.148 (0.142--0.153) & 0.148 (0.143--0.154) & 0.146 (0.140--0.151) & 0.148 (0.142--0.154) & 0.147 (0.141--0.153) \\
\bottomrule
\end{tabular}
\end{table*}

\begin{table*}[]
\centering
\caption{Comparison of total training time and completed epochs reached for different sequence model configurations (context length $C$, model size $S$) in pre-training and fine-tuning setting, with measurements based on the prediction of one-year mortality after the initial \gls{hf} diagnosis.}
\label{stab:train_time}
    \begin{subtable}{\textwidth}
    \centering
    \caption{Pre-training setting (epoch limit: 30).}
    \setlength{\tabcolsep}{4pt}
    \begin{tabular}{@{}llllllllllllll@{}}
    \toprule
     &  & \multicolumn{6}{c}{Total training time in min} & \multicolumn{6}{c}{Completed epochs} \\ \midrule
    $C$ & $S$ & BERT & MBERT & XLNet & Llama & Mamba & Mamba2 & BERT & MBERT & XLNet & Llama & Mamba & Mamba2 \\
    128 & Tiny & 14.7 & 14.9 & 19.6 & 18.2 & 12.9 & 35.3 & 30 & 30 & 30 & 30 & 30 & 30 \\
    128 & Small & 21.5 & 23.8 & 33.9 & 24.7 & 18.7 & 29.4 & 30 & 30 & 30 & 30 & 19 & 19 \\
    128 & Medium & 29.5 & 29.7 & 41.2 & 34.7 & 24.7 & 40.1 & 30 & 30 & 30 & 30 & 16 & 15 \\
    256 & Tiny & 16.4 & 20.2 & 23.6 & 19.7 & 15.3 & 23.0 & 30 & 30 & 30 & 30 & 30 & 30 \\
    256 & Small & 25.5 & 34.0 & 44.3 & 27.6 & 26.0 & 36.4 & 30 & 30 & 30 & 30 & 24 & 21 \\
    256 & Medium & 35.0 & 35.2 & 55.6 & 40.1 & 32.9 & 49.1 & 30 & 30 & 30 & 30 & 18 & 17 \\
    512 & Tiny & 18.8 & 25.3 & 39.9 & 23.5 & 19.2 & 27.7 & 30 & 30 & 30 & 30 & 30 & 30 \\
    512 & Small & 34.5 & 38.3 & 70.6 & 37.1 & 39.4 & 50.0 & 30 & 30 & 30 & 30 & 27 & 25 \\
    512 & Medium & 45.4 & 40.6 & 103.8 & 45.6 & 41.7 & 52.4 & 30 & 30 & 30 & 30 & 20 & 18 \\
    128 & Tiny & 12.4 & 12.7 & 15.5 & 13.7 & 11.7 & 18.3 & 30 & 30 & 30 & 30 & 30 & 30 \\
    128 & Small & 18.9 & 18.2 & 25.6 & 21.2 & 15.8 & 24.4 & 30 & 30 & 30 & 30 & 21 & 18 \\
    128 & Medium & 25.6 & 27.7 & 37.1 & 28.9 & 20.0 & 37.8 & 30 & 30 & 30 & 30 & 16 & 15 \\
    256 & Tiny & 13.8 & 15.0 & 19.0 & 15.1 & 13.4 & 19.9 & 30 & 30 & 30 & 30 & 30 & 30 \\
    256 & Small & 21.2 & 21.0 & 34.5 & 23.8 & 21.4 & 32.5 & 30 & 30 & 30 & 30 & 23 & 22 \\
    256 & Medium & 30.0 & 34.7 & 47.9 & 33.1 & 28.6 & 42.2 & 30 & 30 & 30 & 30 & 18 & 16 \\
    512 & Tiny & 16.9 & 19.6 & 27.7 & 20.3 & 15.7 & 23.1 & 30 & 30 & 30 & 30 & 30 & 30 \\
    512 & Small & 29.9 & 28.2 & 60.0 & 32.1 & 36.5 & 46.1 & 30 & 30 & 30 & 30 & 29 & 26 \\
    512 & Medium & 46.0 & 46.1 & 105.2 & 46.5 & 40.0 & 48.1 & 30 & 30 & 30 & 30 & 19 & 17 \\ \bottomrule
    \end{tabular}
    \end{subtable}
    
    \vspace{0.8em}
    
    \begin{subtable}{\textwidth}
    \centering
    \caption{Fine-tuning setting (epoch limit: 10).}
    \setlength{\tabcolsep}{4pt}
    \begin{tabular}{@{}llllllllllllll@{}}
    \toprule
     &  & \multicolumn{6}{c}{Total training time in min} & \multicolumn{6}{c}{Completed epochs} \\ \midrule
    $C$ & $S$ & BERT & MBERT & XLNet & Llama & Mamba & Mamba2 & BERT & MBERT & XLNet & Llama & Mamba & Mamba2 \\
    128 & Tiny & 2.4 & 3.7 & 4.1 & 3.3 & 3.0 & 4.7 & 5 & 6 & 6 & 5 & 6 & 6 \\
    128 & Small & 5.3 & 4.8 & 4.2 & 3.6 & 4.0 & 15.7 & 7 & 6 & 4 & 4 & 4 & 10 \\
    128 & Medium & 5.1 & 9.5 & 10.6 & 5.8 & 5.3 & 13.4 & 5 & 5 & 7 & 5 & 4 & 5 \\
    256 & Tiny & 2.7 & 2.4 & 4.9 & 3.0 & 3.1 & 4.6 & 6 & 3 & 6 & 5 & 6 & 6 \\
    256 & Small & 8.6 & 5.4 & 9.9 & 4.1 & 5.0 & 15.4 & 10 & 6 & 6 & 4 & 5 & 10 \\
    256 & Medium & 6.8 & 6.1 & 12.7 & 6.7 & 7.4 & 11.4 & 6 & 5 & 7 & 5 & 4 & 4 \\
    512 & Tiny & 3.9 & 5.8 & 9.3 & 2.9 & 3.9 & 5.7 & 6 & 6 & 7 & 4 & 6 & 6 \\
    512 & Small & 7.8 & 6.1 & 14.9 & 5.5 & 5.8 & 18.6 & 7 & 6 & 6 & 5 & 4 & 10 \\
    512 & Medium & 11.6 & 6.9 & 13.8 & 7.6 & 8.1 & 11.1 & 8 & 5 & 4 & 5 & 4 & 4 \\
    128 & Tiny & 2.3 & 3.2 & 4.5 & 2.5 & 2.4 & 3.5 & 6 & 7 & 8 & 5 & 6 & 6 \\
    128 & Small & 3.8 & 2.6 & 5.9 & 4.3 & 3.2 & 13.7 & 6 & 4 & 7 & 6 & 4 & 10 \\
    128 & Medium & 4.2 & 5.1 & 8.5 & 3.9 & 6.4 & 10.0 & 5 & 5 & 7 & 4 & 5 & 4 \\
    256 & Tiny & 4.3 & 2.7 & 5.3 & 2.8 & 2.6 & 4.0 & 10 & 5 & 8 & 5 & 6 & 6 \\
    256 & Small & 3.0 & 5.1 & 4.6 & 4.8 & 3.7 & 14.7 & 4 & 7 & 4 & 6 & 4 & 10 \\
    256 & Medium & 4.8 & 6.9 & 11.1 & 4.5 & 7.7 & 10.5 & 5 & 6 & 7 & 4 & 5 & 4 \\
    512 & Tiny & 5.2 & 5.9 & 8.3 & 3.4 & 3.0 & 3.7 & 10 & 9 & 8 & 5 & 6 & 5 \\
    512 & Small & 3.8 & 3.9 & 13.8 & 6.3 & 6.0 & 17.5 & 4 & 4 & 7 & 6 & 5 & 10 \\
    512 & Medium & 7.4 & 8.5 & 17.4 & 7.7 & 10.2 & 11.2 & 5 & 6 & 5 & 5 & 5 & 4 \\ \bottomrule
    \end{tabular}
    \end{subtable}
\end{table*}

\begin{table*}[]
\centering
\caption{Descriptive statistics and prevalence of clinical prediction tasks of each hospital H for two different internal hospital validation strategies.}
\label{stab:val_hosp_stats}
    \begin{subtable}{\textwidth}
    \setlength{\tabcolsep}{3pt}
    \centering
    \caption{Single-hospital validation.}
\begin{tabular}{@{}lp{2.25cm}p{2.25cm}p{2.25cm}p{2.25cm}p{2.25cm}p{2.25cm}@{}}
    \toprule
    Information & H1 & H2 & H3 & H4 & H5 & H6 \\ \midrule
    $N$ & $3{,}279$ & $9{,}028$ & $15{,}852$ & $5{,}325$ & $1{,}803$ & $7{,}533$ \\
    Age (right-sided) & IQR: 81 (74, 87), \makecell[l]{min: 18, max: 101} & IQR: 81 (73, 87), \makecell[l]{min: 18, max: 105} & IQR: 79 (70, 86), \makecell[l]{min: 18, max: 106} & IQR: 81 (73, 87), \makecell[l]{min: 18, max: 107} & IQR: 81 (73, 87), \makecell[l]{min: 18, max: 106} & IQR: 81 (73, 87), \makecell[l]{min: 18, max: 105} \\
    Sex (\% women) & 44.0\% & 45.4\% & 45.5\% & 46.3\% & 43.8\% & 45.6\% \\
    Hospitalizations & IQR: 2 (1, 3), \makecell[l]{min: 1, max: 32} & IQR: 1 (1, 3), \makecell[l]{min: 1, max: 41} & IQR: 1 (1, 3), \makecell[l]{min: 1, max: 61} & IQR: 2 (1, 3), \makecell[l]{min: 1, max: 39} & IQR: 1 (1, 3), \makecell[l]{min: 1, max: 42} & IQR: 2 (1, 3), \makecell[l]{min: 1, max: 47} \\
    All events & IQR: 146 (64, 328), \makecell[l]{min: 5, max: 3306} & IQR: 146 (67, 310), \makecell[l]{min: 2, max: 4471} & IQR: 144 (61, 313), \makecell[l]{min: 1, max: 9552} & IQR: 175 (86, 361), \makecell[l]{min: 2, max: 5747} & IQR: 148 (66, 322), \makecell[l]{min: 2, max: 4659} & IQR: 128 (60, 274), \makecell[l]{min: 2, max: 8974} \\
    DX events & IQR: 10 (6, 19), \makecell[l]{min: 1, max: 213} & IQR: 10 (6, 19), \makecell[l]{min: 1, max: 265} & IQR: 8 (5, 16), \makecell[l]{min: 1, max: 290} & IQR: 7 (4, 14), \makecell[l]{min: 1, max: 121} & IQR: 8 (5, 16), \makecell[l]{min: 1, max: 172} & IQR: 8 (5, 15), \makecell[l]{min: 1, max: 311} \\
    VIT events & IQR: 31 (0, 97), \makecell[l]{min: 0, max: 1192} & IQR: 26 (0, 79), \makecell[l]{min: 0, max: 1457} & IQR: 28 (0, 89), \makecell[l]{min: 0, max: 2634} & IQR: 43 (0, 115), \makecell[l]{min: 0, max: 2368} & IQR: 45 (0, 110), \makecell[l]{min: 0, max: 1788} & IQR: 30 (0, 83), \makecell[l]{min: 0, max: 2337} \\
\bottomrule
    \end{tabular}
    \end{subtable}
    
    \vspace{0.8em}
    
    \begin{subtable}{\textwidth}
    \centering
    \caption{Cross-hospital validation.}
    \setlength{\tabcolsep}{3pt}
    \begin{tabular}{@{}lp{2.25cm}p{2.25cm}p{2.25cm}p{2.25cm}@{}}
    \toprule
    Information & All & H1--H4 & H5 & H6 \\ 
    & & (Development set) & (Test set 1) &  (Test set 2) \\ \midrule
    $N$ & $42{,}820$ & $33{,}484$ & $1{,}803$ & $7{,}533$ \\
    Age (right-sided) & IQR: 80 (72, 87), \makecell[l]{min: 18, max: 107} & IQR: 81 (73, 87), \makecell[l]{min: 18, max: 107} & IQR: 81 (73, 87), \makecell[l]{min: 18, max: 106} & IQR: 81 (73, 87), \makecell[l]{min: 18, max: 105} \\
    Sex (\% women) & 45.4\% & 45.5\% & 43.8\% & 45.6\%  \\
    Hospitalizations & IQR: 2 (1, 3), \makecell[l]{min: 1, max: 61} & IQR: 2 (1, 3), \makecell[l]{min: 1, max: 61} & IQR: 1 (1, 3), \makecell[l]{min: 1, max: 42} & IQR: 2 (1, 3), \makecell[l]{min: 1, max: 47} \\
    All events & IQR: 145 (65, 312), \makecell[l]{min: 1, max: 9552} & IQR: 149 (67, 321), \makecell[l]{min: 1, max: 9552} & IQR: 148 (66, 322), \makecell[l]{min: 2, max: 4659} & IQR: 128 (60, 274), \makecell[l]{min: 2, max: 8974} \\
    DX events & IQR: 9 (5, 16), \makecell[l]{min: 1, max: 311} & IQR: 9 (5, 17), \makecell[l]{min: 1, max: 290} & IQR: 8 (5, 16), \makecell[l]{min: 1, max: 172} & IQR: 8 (5, 15), \makecell[l]{min: 1, max: 311} \\
\bottomrule
    \end{tabular}
    \end{subtable}
\end{table*}

\begin{table*}[!ht]
\centering
\caption{Evaluation of singe-hospital validation. Each model was trained with vocabulary \voctwo{}, $C=512$, $\mathrm{Medium}$ model size. \textbf{Best} and \underline{second best} task-specific performances for each hospital H and the average across hospitals for each model are highlighted in bold or underlined.}
\label{stab:eval_shv}
    \begin{subtable}{\textwidth}
    \caption{Evaluation of \gls{auprc} ($\uparrow$).}
    \label{stab:eval_shv_auprc}
    \setlength{\tabcolsep}{3pt}
    \begin{tabular}{@{}llllllll@{}}
    \toprule
    H & XGBoost & BERT & XLNet & MBERT & Llama & Mamba & Mamba2 \\ \midrule
    \multicolumn{8}{c}{\tone{}} \\
    Avg. & 0.506$\pm$0.035 & \underline{0.511}$\pm$0.038 & 0.504$\pm$0.042 & 0.505$\pm$0.041 & \textbf{0.516}$\pm$0.043 & 0.492$\pm$0.042 & 0.503$\pm$0.033 \\
    1 & \textbf{0.517} (0.448--0.584) & 0.504 (0.439--0.574) & \underline{0.509} (0.445--0.580) & 0.488 (0.419--0.558) & 0.480 (0.414--0.547) & 0.484 (0.416--0.555) & 0.471 (0.401--0.545) \\
    2 & 0.505 (0.462--0.547) & 0.519 (0.476--0.561) & 0.513 (0.472--0.558) & \underline{0.522} (0.481--0.568) & \textbf{0.553} (0.509--0.597) & 0.507 (0.462--0.551) & \underline{0.522} (0.480--0.567) \\
    3 & 0.535 (0.499--0.567) & 0.532 (0.495--0.569) & 0.524 (0.491--0.558) & \textbf{0.536} (0.504--0.572) & \textbf{0.536} (0.501--0.574) & 0.526 (0.491--0.562) & 0.531 (0.496--0.567) \\
    4 & 0.515 (0.461--0.574) & \textbf{0.554} (0.496--0.610) & 0.542 (0.486--0.597) & 0.544 (0.487--0.600) & \underline{0.550} (0.492--0.607) & 0.535 (0.477--0.596) & 0.540 (0.481--0.599) \\
    5 & 0.438 (0.348--0.532) & 0.441 (0.353--0.531) & 0.422 (0.338--0.515) & 0.431 (0.348--0.523) & \underline{0.446} (0.360--0.541) & 0.417 (0.345--0.501) & \textbf{0.463} (0.367--0.568) \\
\bottomrule
    \end{tabular}
    \end{subtable}

    \vspace{0.8em}

    \begin{subtable}{\textwidth}
    \centering
    \caption{Evaluation of \gls{auroc} ($\uparrow$).}
    \label{stab:eval_shv_auroc}
    \setlength{\tabcolsep}{3pt}
    \begin{tabular}{@{}llllllll@{}}
    \toprule
    H & XGBoost & BERT & XLNet & MBERT & Llama & Mamba & Mamba2 \\ \midrule
    \multicolumn{8}{c}{\tone{}} \\
    Avg. & 0.616$\pm$0.034 & \underline{0.622}$\pm$0.041 & 0.617$\pm$0.039 & 0.621$\pm$0.048 & \textbf{0.635}$\pm$0.037 & 0.620$\pm$0.037 & 0.620$\pm$0.038 \\
    1 & \textbf{0.614} (0.561--0.663) & 0.605 (0.556--0.656) & \underline{0.613} (0.567--0.663) & 0.576 (0.526--0.629) & 0.597 (0.546--0.647) & 0.590 (0.541--0.642) & 0.582 (0.532--0.630) \\
    2 & 0.611 (0.579--0.644) & 0.641 (0.609--0.671) & 0.630 (0.599--0.659) & 0.643 (0.612--0.673) & \textbf{0.666} (0.636--0.694) & 0.640 (0.611--0.668) & \underline{0.646} (0.616--0.673) \\
    3 & 0.654 (0.631--0.677) & 0.652 (0.630--0.673) & 0.650 (0.629--0.673) & 0.665 (0.643--0.689) & 0.658 (0.637--0.680) & \textbf{0.667} (0.645--0.689) & \underline{0.666} (0.645--0.689) \\
    4 & 0.619 (0.578--0.658) & \underline{0.655} (0.617--0.693) & 0.641 (0.602--0.679) & 0.653 (0.615--0.692) & \textbf{0.658} (0.620--0.695) & 0.645 (0.608--0.683) & 0.643 (0.604--0.679) \\
    5 & 0.555 (0.484--0.626) & 0.546 (0.474--0.620) & 0.541 (0.464--0.614) & 0.546 (0.471--0.618) & \textbf{0.579} (0.509--0.648) & 0.569 (0.498--0.644) & \underline{0.574} (0.502--0.646) \\
\bottomrule
    \end{tabular}
    \end{subtable}

    \vspace{0.8em}
    
    \begin{subtable}{\textwidth}
    \centering
    \caption{Evaluation of Brier score ($\downarrow$).}
    \label{stab:eval_shv_brier}
    \setlength{\tabcolsep}{3pt}
    \begin{tabular}{@{}llllllll@{}}
    \toprule
    H & XGBoost & BERT & XLNet & MBERT & Llama & Mamba & Mamba2 \\ \midrule
    \multicolumn{8}{c}{\tone{}} \\
    Avg. & 0.251$\pm$0.021 & \underline{0.230}$\pm$0.008 & \underline{0.230}$\pm$0.006 & 0.232$\pm$0.007 & \textbf{0.229}$\pm$0.007 & 0.233$\pm$0.01 & 0.232$\pm$0.006 \\
    1 & 0.256 (0.234--0.279) & 0.239 (0.225--0.254) & \textbf{0.232} (0.221--0.243) & 0.238 (0.228--0.249) & 0.236 (0.223--0.249) & \underline{0.235} (0.224--0.246) & 0.237 (0.224--0.251) \\
    2 & 0.245 (0.233--0.256) & 0.228 (0.220--0.235) & 0.229 (0.223--0.235) & 0.231 (0.227--0.235) & \textbf{0.224} (0.213--0.233) & 0.228 (0.221--0.235) & \underline{0.227} (0.220--0.233) \\
    3 & 0.225 (0.217--0.233) & \underline{0.221} (0.215--0.228) & \underline{0.221} (0.215--0.227) & 0.222 (0.214--0.230) & \underline{0.221} (0.215--0.227) & \textbf{0.220} (0.212--0.227) & 0.222 (0.214--0.230) \\
    4 & 0.253 (0.237--0.270) & \textbf{0.226} (0.215--0.236) & 0.228 (0.219--0.235) & 0.229 (0.221--0.237) & \textbf{0.226} (0.215--0.236) & 0.229 (0.216--0.239) & 0.230 (0.219--0.242) \\
    5 & 0.288 (0.253--0.323) & 0.240 (0.227--0.254) & 0.241 (0.225--0.256) & 0.242 (0.222--0.262) & \underline{0.238} (0.216--0.261) & 0.247 (0.223--0.272) & \textbf{0.237} (0.225--0.249) \\
\bottomrule
    \end{tabular}
    \end{subtable}
\end{table*}

\begin{table*}[h]
\centering
\caption{Evaluation of cross-hospital validation. \textbf{Best} and \underline{second best} task-specific \gls{auprc} ($\uparrow$) performances are highlighted in bold or underlined across hospital stratifications. \textsuperscript{\textdagger}\gls{tgds} configuration..}
\label{stab:eval_chv_auprc}
\begin{tabular}{llllllll}
\toprule
Model & $C$ & $S$ & Vocabulary & History & H1--H4 & H5 & H6 \\ \midrule
\multicolumn{8}{c}{\tone{}} \\
\multicolumn{8}{l}{\chvvarone{}} \\
BERT & 512 & Medium & \voctwo{} & \histcontext{} & 0.533 (0.508--0.556) & 0.523 (0.485--0.561) & 0.545 (0.527--0.565) \\
XLNet & 512 & Medium & \voctwo{} & \histcontext{} & 0.532 (0.508--0.556) & 0.526 (0.487--0.563) & 0.535 (0.516--0.553) \\
MBERT & 512 & Medium & \voctwo{} & \histcontext{} & 0.524 (0.499--0.546) & \underline{0.537} (0.498--0.578) & 0.544 (0.526--0.563) \\
Llama\textsuperscript{\textdagger} & 512 & Medium & \voctwo{} & \histcontext{} & 0.540 (0.516--0.563) & 0.524 (0.486--0.561) & 0.553 (0.534--0.572) \\
Mamba & 512 & Medium & \voctwo{} & \histcontext{} & 0.537 (0.513--0.560) & 0.524 (0.484--0.563) & 0.543 (0.524--0.563) \\
\bottomrule
\end{tabular}
\end{table*}

\begin{table*}[]
\centering
\caption{Evaluation of cross-hospital validation. \textbf{Best} and \underline{second best} task-specific \gls{auroc} ($\uparrow$) performances are highlighted in bold or underlined across hospital stratifications. \textsuperscript{\textdagger}\gls{tgds} configuration.}
\label{stab:eval_chv_auroc}
\begin{tabular}{llllllll}
\toprule
Model & $C$ & $S$ & Vocabulary & History & H1--H4 & H5 & H6 \\ \midrule
\multicolumn{8}{c}{\tone{}} \\
\multicolumn{8}{l}{\chvvarone{}} \\
BERT & 512 & Medium & \voctwo{} & \histcontext{} & 0.658 (0.642--0.673) & 0.653 (0.628--0.679) & 0.661 (0.648--0.673) \\
XLNet & 512 & Medium & \voctwo{} & \histcontext{} & 0.656 (0.641--0.671) & 0.655 (0.629--0.680) & 0.650 (0.637--0.662) \\
MBERT & 512 & Medium & \voctwo{} & \histcontext{} & 0.653 (0.637--0.669) & \underline{0.665} (0.640--0.690) & 0.662 (0.649--0.674) \\
Llama\textsuperscript{\textdagger} & 512 & Medium & \voctwo{} & \histcontext{} & 0.666 (0.651--0.680) & 0.661 (0.635--0.686) & 0.667 (0.654--0.679) \\
Mamba & 512 & Medium & \voctwo{} & \histcontext{} & 0.659 (0.645--0.675) & 0.658 (0.631--0.683) & 0.664 (0.651--0.676) \\
\bottomrule
\end{tabular}
\end{table*}

\begin{table*}[]
\centering
\caption{Evaluation of cross-hospital validation. \textbf{Best} and \underline{second best} task-specific Brier score ($\downarrow$) performances are highlighted in bold or underlined across hospital stratifications. \textsuperscript{\textdagger}\gls{tgds} configuration.}
\label{stab:eval_chv_brier}
\begin{tabular}{llllllll}
\toprule
Model & $C$ & $S$ & Vocabulary & History & H1--H4 & H5 & H6 \\ \midrule
\multicolumn{8}{c}{\tone{}} \\
\multicolumn{8}{l}{\chvvarone{}} \\
BERT & 512 & Medium & \voctwo{} & \histcontext{} & 0.222 (0.218--0.225) & 0.223 (0.216--0.229) & 0.222 (0.219--0.225) \\
XLNet & 512 & Medium & \voctwo{} & \histcontext{} & 0.224 (0.219--0.229) & 0.223 (0.216--0.231) & 0.227 (0.223--0.230) \\
MBERT & 512 & Medium & \voctwo{} & \histcontext{} & 0.226 (0.221--0.231) & 0.222 (0.214--0.230) & 0.224 (0.220--0.228) \\
Llama\textsuperscript{\textdagger} & 512 & Medium & \voctwo{} & \histcontext{} & 0.224 (0.219--0.229) & 0.224 (0.216--0.232) & 0.224 (0.220--0.228) \\
Mamba & 512 & Medium & \voctwo{} & \histcontext{} & 0.222 (0.217--0.227) & 0.223 (0.215--0.231) & 0.222 (0.219--0.226) \\
\bottomrule
\end{tabular}
\end{table*}

\end{document}